\documentclass[twoside]{article}

\usepackage[accepted]{aistats2024}

\usepackage{graphicx}
\usepackage{arydshln}

\usepackage[utf8]{inputenc} %
\usepackage[T1]{fontenc}    %
\usepackage{hyperref}       %
\usepackage{url}            %
\usepackage{booktabs}       %
\usepackage{amsfonts}       %
\usepackage{nicefrac}       %
\usepackage{microtype}      %
\usepackage{bbm}
\usepackage{xcolor}         %
\usepackage{subcaption}
\usepackage{multirow}
\usepackage{tikz}
\usepackage{colortbl}

\usepackage[round]{natbib}

\usepackage{arydshln}
\usepackage{enumitem}
\usepackage{amsmath}
\usepackage{amsthm}
\usepackage{bbm}
\usepackage{amssymb}
\usepackage{subcaption}
\usepackage{dsfont}
\usepackage{layouts}
\usepackage{float}

\usepackage[capitalise]{cleveref}

\DeclareMathAlphabet\mathbfcal{OMS}{cmsy}{b}{n}

\definecolor{dot_red}{HTML}{ff0000}
\definecolor{dot_green}{HTML}{05AB00}
\definecolor{dot_blue}{HTML}{0077FF}
\definecolor{dot_orange}{HTML}{FF9E00}
\newcommand{\mycirc}[1][black]{\Large\textcolor{#1}{\ensuremath\bullet}}

\newcommand{\E}{\mathbb{E}}
\newcommand{\hist}{\mathcal{H}}

\newcommand{\Prob}{p}

\newcommand{\prob}{\Prob}

\newcommand{\ind}{\mathds{1}}%
\newcommand{\sep}{\!\;|\;\!}
\newcommand{\hit}{\text{hit}}
\newcommand{\vocab}{\mathcal{X}}
\newcommand{\set}{X}
\newcommand{\itemnum}{K}

\newcommand{\gmidrule}{
  \arrayrulecolor{lightgray}
  \specialrule{\lightrulewidth}{0.4\aboverulesep}{0.6\belowrulesep}
  \arrayrulecolor{black}
}

\begin{document}

\twocolumn[

\aistatstitle{Probabilistic Modeling for Sequences of Sets in Continuous-Time}

\aistatsauthor{ Yuxin Chang${}^1$ \And Alex Boyd${}^2$ \And  Padhraic Smyth${}^{1,2}$ }
\runningauthor{Yuxin Chang, Alex Boyd, Padhraic Smyth}
\aistatsaddress{${}^1$Department of Computer Science \quad ${}^2$Department of Statistics \\ 
University of California, Irvine} ]

\begin{abstract}
    Neural marked temporal point processes have been a valuable addition to the existing toolbox of statistical parametric models for continuous-time event data. These models are useful for sequences where each event is associated with a single item (a single type of event or a ``mark'')---but such models are not suited for the practical situation where each event is associated with a set of items. In this work, we develop a general framework for modeling set-valued data in continuous-time, compatible with any intensity-based recurrent neural point process model.  In addition, we develop inference methods that can use such models to answer probabilistic queries such as ``the probability of item $A$ being observed before item $B$,'' conditioned on sequence history. Computing exact answers for such queries is generally intractable for neural models due to both the continuous-time nature of the problem setting and the combinatorially-large space of potential outcomes for each event. To address this, we develop a class of importance sampling methods for querying with set-based sequences and demonstrate orders-of-magnitude improvements in efficiency over direct sampling via systematic experiments with four real-world datasets. We also illustrate how to use this framework to perform model selection using likelihoods that do not involve one-step-ahead prediction.

\end{abstract}

\section{INTRODUCTION}

Modeling and prediction of discrete event data in continuous time is of broad interest in machine learning and statistics. Such data are often modeled via marked temporal point processes (MTPPs) where each event is associated with a single mark (or event type) from a finite vocabulary of marks. These models are characterized by \textit{mark-specific} intensities, interpreted as the instantaneous rate of occurrence of each mark. 

MTPP models are useful across a broad range of applications involving complex temporal phenomena, such as disease transmission \citep{rambhatla2022toward,holbrook2022viral,sarita2022,giudici2023network},  neuronal activity \citep{pfaffelhuber2022mean,bonnet2023inference}, and financial event data \citep{bacry2014hawkes,hawkes2018hawkes,wehrli2022classification}.
Early work on MTPPs relied largely on relatively simple parametric modeling approaches \citep{hawkes1971spectra,brillinger1975identification,daley2003introduction}. More recently, there has been significant interest in developing more flexible neural MTPP models, such as recurrent MTPPs \citep{du2016recurrent} and neural Hawkes processes \citep{mei2017neural}.
However, these models were all developed under the assumption that no more than one event of a single type happens at the same time.

\begin{figure}
    \centering
    \includegraphics[width=0.6\columnwidth]{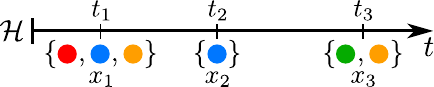}
    \caption{Example sequence of set-valued events, where the item space is $\{\mycirc[dot_red], \mycirc[dot_blue], \mycirc[dot_green], \mycirc[dot_orange]\}$ and $\hist$ refers to history.}
    \label{fig:sample_seq}
\end{figure}

In this paper, we are interested in the situation where each event is associated with a {\it set} of items that occur simultaneously, rather than being associated with a single item (e.g., see \cref{fig:sample_seq}). An example is shopping data, where a set of items are purchased at the time of each purchasing event. %
Given $K$ items, one approach to modeling such data would be to directly apply existing MTPP models by associating each of the possible $2^K$ sets of items with a unique mark
\citep{turkmen2020fastpoint, ma2021bridging}.
However, %
this results in an exponential growth in the number of parameters as a function of the number of items $K$. In addition, this representation does not capture the underlying set structure, making inference and querying difficult. For example,  simple queries such as ``Is item $A$ more likely to be purchased than item $B$ in the next basket?'' could require enumeration over an exponential number of subsets containing relevant items.

To address this, we develop alternative representations for modeling sets directly. Our general approach is compatible with any intensity-based (e.g., black-box) recurrent MTPP model. More specifically, we propose a general framework to model set-valued continuous-time event data based on recurrent MTPP models, where event types are subsets of a finite set. Furthermore, we %
investigate flexible sampling methods to answer general probabilistic queries beyond predicting the next immediate event, defined on (subsets of) items in the context of $2^K$ possible subsets. Example queries include hitting time queries, such as ``What is the probability that any item in the subset $A$ will occur before time $t$,'' and $A$-before-$B$ queries that compute the probability that an item in set $A$ will occur before an item in set $B$.
While there is recent prior work on answering probabilistic queries in the standard MTPP setting \citep{boyd2023probabilistic}, here we study these queries for subsets, where the queries can in general be more complicated. 

Based on systematic experiments on four real-world datasets, we empirically demonstrate that: (i) our proposed models have significantly better predictive power than alternative baselines; (ii) our importance-sampling approach to querying is orders of magnitude more efficient than naive sampling methods; and (iii) our proposed query-based log-likelihood metrics are effective for model selection.

\section{RELATED WORK}
\paragraph{Sequential Set Predictions} 
There has been prior work on modeling sequences of set-valued events 
for problems such as next basket predictions in recommendation systems  \citep{rendle2010factorizing, yu2016dynamic, hu2019sets2sets, shou2023concurrent} and modeling bags of words over time in dynamic topic modeling \citep{wang2006topics,wang2008continuous}.
Their primary focus has been on modeling and predicting the items (products, words, etc.) in each set, conditioned on known times for events.
In contrast, we focus on fully generative probabilistic models where we can make inferences about both time and sets, allowing us to marginalize over uncertainties about intermediate events to answer queries about the future.

\paragraph{Neural MTPPs} 
Marked temporal point processes (MTPPs) are generative models that jointly model sequences of event times and types. Early work on neural MTPPs \citep{du2016recurrent,mei2017neural} has been followed by a burst of activity in this area, developing a variety of subsequent variations of the initial ideas \citep{shchur2021neural}, with extensions to handle missing data \citep{shchur2019intensity, mei2019imputing, gupta2021learning}, Monte Carlo inference techniques \citep{shelton2018hawkes}, long-term forecasting \citep{deshpande2021long, xue2022hypro}, and computationally-scalable methods for large mark vocabularies \citep{turkmen2020fastpoint}. All of this prior work has focused on the classical MTPP framework, where each event is associated with a single type, whereas our approach differs in that we allow multiple types for each event. 

In earlier work, \cite{boyd2023probabilistic} introduced the idea of probabilistic querying with continuous-time neural MTPP models, showing that answering such queries analytically is generally intractable, and demonstrating how importance sampling can be used to provide approximate query answers efficiently. %
We build on this work and demonstrate how to extend the querying frameworks proposed by \cite{boyd2023probabilistic} to leverage the structure of set-valued MTPPs.

\paragraph{Determinantal Point Processes (DPPs)} 
DPPs are probabilistic models that efficiently define a distribution over all $2^K$ subsets of $K$ items, characterized by negative item-to-item correlations and marginal probabilities \citep{macchi1975coincidence, kulesza2012determinantal}.
These models are often appealing in that various conditioning and inference operations involving subsets of items can be carried out in closed form. DPPs have been successfully applied
in various machine learning tasks such as pose estimation \citep{kulesza2010structured} and multi-label classification \citep{xie2017deep}; however, to our knowledge there is no existing work which uses DPPs for sequences of sets, particularly over continuous-time.

\section{MODELING FRAMEWORK}

\begin{figure}
    \centering
    \includegraphics[width=0.9\columnwidth]{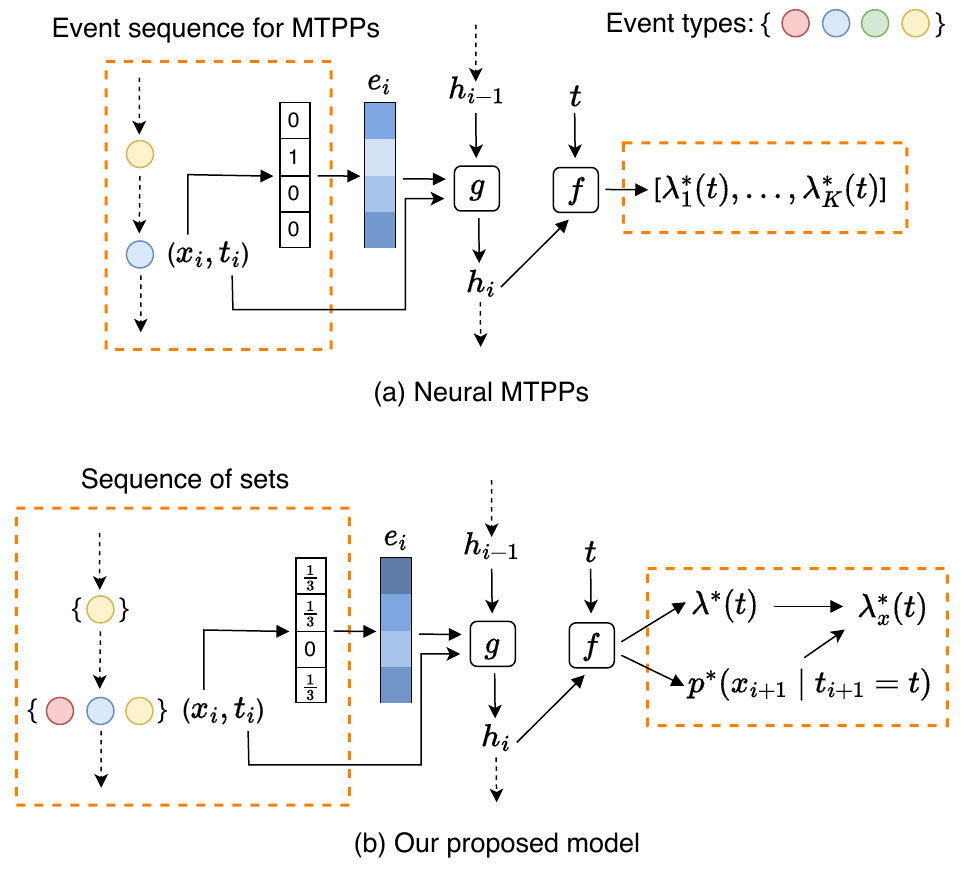}
    \caption{A comparison of (a) traditional neural MTPPs and(b) our proposed model for sequences of sets, where the differences are highlighted in the orange boxes. In (a) $x_i$ represents a single item, whereas in (b) $x_i$ is a set of items.}
    \label{fig:illustration}
\end{figure}

\subsection{Preliminaries}

Let $\tau_i \in \mathbb{R}_{\geq 0}$ be the random variable of the $i$th distinct event time in a continuous-time sequence, i.e., $\forall i: \tau_i < \tau_{i+1}
$, where $i$ is a positive integer, and $t_i$ is its realization. 
Alongside the time of occurrence, each event also possesses an additional piece of information, $X_i \in \vocab$, commonly referred to as a \textit{mark}. 
We use $X_i$ and $x_i$ to denote the corresponding random variable and realization of mark $i$. Define the \textit{history} of the sequence for any time interval ${[a, b] \subset \mathbb{R}_{\geq 0}}$ as 
\begin{align}
\hist[a, b] = \{(\tau_i, \set_i) \mid \forall i \in \mathbb{N}^+, \tau_i \in [a, b] \}.    
\end{align}
The history $\hist(a, b]$ and $\hist[a, b)$ are defined similarly. We use the abbreviation for the history up to, but not including time $t$, as $\hist(t):=\hist[0,t)$.

Marked (or multivariate) temporal point processes (MTPPs) are a class of generative models for sequential data with continuous timestamps and marked information.\footnote{While our treatment of MTPPs assumes that marks are discrete and finite, the general framework allows for continuous-valued marks as well as more complicated domains, e.g., combinations of discrete- and continuous-valued marks.} MTPPs are fully parameterized by non-negative, \textit{mark-specific intensity} functions %
${\lambda^*_x(t) := \lambda_x(t \sep \hist(t))}$ 
which represent the instantaneous rate of occurrence for each event type $x$ at time $t$ conditioned on the entire history over the timespan $[0,t)$. In general, we use ``$*$'' to indicate conditioning on $\hist(t)$.
Summing up all mark-specific intensities yields the \textit{total intensity} $\lambda^*(t) := \sum_{x\in\vocab} \lambda^*_x(t)$,
also known as the \textit{ground intensity}. The conditional mark distribution is then defined as a ratio of intensities: 
${p^*(x\sep t) := p(X_i=x \sep \tau_i=t, \hist(\tau_i)) \equiv \lambda^*_x(t) / \lambda^*(t).}$
The log-likelihood of a particular sequence $\hist(T)$ with $N$ events decomposes into the following form:
\begin{align}
\mathcal{L}(\hist(T))\!=\! \underbrace{\sum_{i=1}^{N} \log p^*\!(x_i\sep t_i)}_{\mathcal{L}_\text{Mark}} \!+\!\underbrace{\sum_{i=1}^N \log \lambda^*\!(t_i) \!-\!\!\int_0^T\!\!\lambda^*\!(s)ds}_{\mathcal{L}_\text{Time}}\!.
\label{eq:logl}
\end{align}
Here, $\mathcal{L}_\text{Time}$ is the log-likelihood of a temporal point process that treats marks as fixed covariates and $\mathcal{L}_\text{Mark}$ is akin to a sequential classification cross-entropy.

In recurrent MTPPs \citep{du2016recurrent,mei2017neural},
hidden states $\mathbf{h}(t)$ from continuous-time recurrent neural networks (RNNs) are used to summarize the history $\hist(t)$ up to time $t$. %
This results in the vector of marked intensities being modeled by some parameterized transformation of the hidden state, e.g., $\lambda_x^*(t) := \lambda_x(t\sep \mathbf{h}(t)):= \exp(\mathbf{u}_x \cdot \mathbf{h}(t)+b_x)$ for $x\in\mathcal{X}$. Hidden states are typically computed recursively:
\begin{align}
\mathbf{h}(t) & := \mathbf{f}(\mathbf{h}_i, t) \text{ for } t_i < t \leq t_{i+1} \\
\mathbf{h}_i & := \mathbf{g}(\mathbf{h}_{i-1}, t_i, \mathbf{e}(x_i)), \label{eq:rnn_input}
\end{align}
where $\mathbf{h}_0$ is a learnable parameter and $\mathbf{e}(x)$ embeds the mark $x$ as a dense vector which we assume to take the form of $\mathbf{e}(x)=\mathbf{w}_x$ for learnable vectors $\mathbf{w}_x$ for $x\in\vocab$.

\subsection{Setting of Interest: Set-Valued Sequences}
We are primarily interested in the scenario where the marks $X_i$ are not just simple discrete labels but possess %
\textit{set} structure, meaning they can be composed of some subset of possible \textit{items}. 
To avoid potential confusion, we will typically refer to ``marks'' as ``sets'' to differentiate from the items contained within (e.g., $\mathcal{L}_\text{Mark}$ is now $\mathcal{L}_\text{Set}$).
The possible items are assumed to be a fixed set of $\itemnum$ unique values, with the range of possible sets spanning $\vocab:=\mathcal{P}(\{1, \dots, \itemnum\})$, which is the power set of the set $\{1, \dots, \itemnum\}$ and includes the empty set.

As mentioned earlier, while it is possible
to naively map each possible set in $\vocab$ to a unique, discrete label, 
this does not take advantage of the inherent structure present in the set-valued marks. For example, this naive mapping would treat sets $\{a, b, c\}$ and $\{a, b\}$ as completely separate values, ignoring the fact that they share items $a$ and $b$
and could lead to an inordinate amount of parameters, i.e., $|\mathcal{X}|=2^K$.

Our work aims to take advantage of the structured nature of the sets $X_i$ to more efficiently and effectively model an MTPP for set-valued marks. In addition, we develop a general approach for set-valued marks so that it allows any intensity-based recurrent MTPP model to be used within our proposed framework. \cref{fig:illustration} provides a high-level illustration of our proposed model and how it differs from typical neural MTPPs.

\subsection{Modeling Approach}

\paragraph{Set Representation and Intake} 
To adapt an existing recurrent MTPP to set-valued marks, we must first define a method of using sets as inputs when calculating the hidden state. Put differently, we need to embed a set into a dense vector representation in order to make it compatible for $\mathbf{g}$ in \cref{eq:rnn_input}. We propose embedding the individual items within a given set $x$, and then representing the composition of them additively:
\begin{align}
\mathbf{e}(x) := \begin{cases}
    \frac{1}{|x|}\sum_{k \in x} \mathbf{w}_k & \text{if } x \neq \emptyset \\
    \mathbf{0} & \text{if } x = \emptyset
\end{cases} \label{eq:embed}
\end{align}
where $\{\mathbf{w}_k\}_{k=1}^K$ are learned. We normalize by the number of items in the set to keep embeddings of different sets on roughly the same scale.

With the embedding of sets adapted to the same format as typical embedded marks for the original neural MTPPs, $\mathbf{h}(t)$ can be computed in the usual manner using the same $\mathbf{f}$ and $\mathbf{g}$ functions for a given base recurrent MTPP model. The parameters of $\mathbf{f}$ and $\mathbf{g}$, along with $\{\mathbf{w}_k\}_{k=1}^K$, will be denoted collectively as $\theta$, with hidden states written as $\mathbf{h}_\theta(t)$ to emphasize this parameterization.

\paragraph{Intensity Modeling}
Since the total number of possible sets is $2^K$, it becomes intractable as $K$ grows to model $2^K$ different marked intensity values, each with learnable parameters.
Instead, we choose to separate the distribution over time from the distribution over sets by noting the following property:
\begin{align}
\lambda_x^*(t) := \lambda^*(t)p^*(x \sep t). \label{eq:factor}
\end{align}
As such, it is sufficient to separately model the total intensity function and the conditional mark distribution over sets, where the latter is discussed further in the sections below on set modeling.

All recurrent MTPPs support directly modeling the total intensity of a process by assuming %
there is only a single possible mark value. 
For example, the neural Hawkes model computes a marked intensity function via ${\lambda^*_x(t) = s_x \log ( 1 + \exp(\mathbf{u}_x \cdot \mathbf{h}(t) / s_x))}$ with mark-specific parameters $s_x$ and $\mathbf{u}_x$. We adapt this to model the total intensity by having what were original mark-specific parameters now globally shared: $\lambda^*(t) = s \log ( 1 + \exp(\mathbf{u} \cdot \mathbf{h}(t) / s))$. All parameters that are solely used to compute $\lambda$ from the hidden state $\mathbf{h}(t)$ (such as $s$ and $\mathbf{u}$ from the previous example) will be referred to as $\phi$, with the total intensity written as $\lambda^*_\phi(t)$ to denote this parameterization. 

The modeling decisions made thus far allow us to define the temporal component of the log-likelihood of a sequence $\hist(T)$ with $N$ events as follows:
\begin{align}
\mathcal{L}&_\text{Time}(\theta, \phi; \hist(T)) \notag \\
& := \sum_{i=1}^N \log \lambda_\phi(t_i \sep \mathbf{h}_\theta(t_i)) - \int_0^T \lambda_\phi(s \sep \mathbf{h}_\theta(s))ds.
\end{align}

All that remains for our proposed framework is to determine the conditional set distribution ${p^*(x \sep t) := p(x \sep t, \mathbf{h}_\theta(t))}$. We present two different approaches to this below.

\paragraph{Set Modeling: Dynamic Bernoulli}
Our first approach to set modeling will be referred to as a Dynamic Bernoulli model.
Specifically, given a hidden state $\mathbf{h}_\theta(t)$, the conditional set distribution of the next set $X$ is parameterized via
\begin{align}
p(X=x \sep t, \mathbf{h}(t)) & := \prod_{k=1}^K \rho_k(t)^{\ind(k \in x)}(1-\rho_k(t))^{\ind(k \notin x)} \notag\\
\rho_k(t) & := \sigma(\mathbf{v}_k \cdot \mathbf{n}(\mathbf{h}(t)) + b_k), \label{eq:intensity2logits}
\end{align}
where $\ind(\cdot)$ is the indicator function, $\sigma(\cdot)$ is the sigmoid function, $\mathbf{n}$ is a feed-forward neural network, and $\mathbf{v}_k$ and $b_k$ are learnable parameters for $k=1,\dots,K$. The values $\rho_k(t)$ can be interpreted as the probability of item $k$ appearing in the set $X$ at time $t$. This model assumes that the presence of each item is conditionally independent,  \emph{conditioned on the history $\mathbf{h}(t)$}. It should be noted that the model allows for there to be significant marginal correlation between items in a set, especially considering the flexibility of the feedforward network $\mathbf{n}$. More formally, we assume $(k \in X_i \perp k' \in X_i) \sep \tau_i, \hist[0,\tau_{i-1}]$, but marginally, it follows that $(k \in X_i \not\perp k' \in X_i) \sep \hist[0,\tau_{i-1}]$ for $k\neq k' \in \{1,\dots,K\}$ due to how $\mathbf{h}(t)$ evolves over time.

All of the parameters of $\mathbf{n}$ as well as  $\mathbf{v}_k$ and $b_k$ for all $k$ will be referred to as $\omega$, with the conditional set distribution written as $p^*_\omega(x\sep t)$. The set-specific contribution to the log-likelihood of the Dynamic Bernoulli model is defined in the following manner:
\begin{align}
&\mathcal{L}_\text{Set}(\omega, \theta; \hist(T)) := \sum_{i=1}^N \log p_\omega(x_i \sep t_i, \mathbf{h}_\theta(t_i)) \\
& = \!\sum_{i=1}^N \sum_{k=1}^K \ind(k \!\in\! x_i)\log \rho_k(t_i) \!+\! \ind(k \!\notin\! x_i) \log (1 \!-\! \rho_k(t_i)).\notag
\end{align}

\paragraph{Set Modeling: Dynamic DPPs} A potential limitation of the Dynamic Bernoulli approach is that correlations between items are not modeled beyond conditioning on hidden states $\mathbf{h}(t)$. An instance where this might matter is when the presence of one item actively inhibits the other in a set, but both are marginally likely to occur conditioned on the history.

With this in mind we can model a more expressive set distribution by utilizing determinantal point processes (DPPs) \citep{kulesza2012determinantal}. DPPs are a class of distributions over sets that allow for modeling marginal likelihoods of item inclusions as well as negative pair-wise correlations. They can be fully parameterized by a $K\times K$ symmetric and positive semidefinite matrix $\mathbf{L}$. Namely, $p(X=x) = \frac{\text{det}(\mathbf{L}_x)}{\text{det}(\mathbf{L} + \mathbf{I})}$, where $\mathbf{L}_x = (\mathbf{L}_{ij})_{i,j\in x} \in \mathbb{R}^{|x| \times |x|}$ and $\mathbf{I}$ is the $\itemnum \times \itemnum$ identity matrix. 

One way to allow a DPP to condition on the history up to time $t$ (and thus allow it to be ``dynamic'') is to parameterize the $\mathbf{L}$ matrix as a function of the hidden state $\mathbf{h}(t)$. We consider the following parameterization:
\begin{align}
p_\omega(x \sep t, \mathbf{h}_\theta(t)) & := \frac{\text{det}(\mathbf{L}_x(t))}{\text{det}(\mathbf{L}(t) + \mathbf{I})} \\
\mathbf{L}_x(t) & := (\mathbf{L}_{ij}(t))_{i,j\in x} \\
\mathbf{L}_{ij}(t) & := \mathbf{n}_i(\mathbf{h}(t))\frac{\mathbf{w}_i\cdot\mathbf{w}_j}{\lVert\mathbf{w}_i\rVert\lVert\mathbf{w}_j\rVert}\mathbf{n}_j(\mathbf{h}(t))
\end{align}
where $\mathbf{n}: \mathbb{R}^d \to \mathbb{R}^{K}$
is a feedforward neural network with parameters $\omega$ and $\mathbf{w}_i$ and $\mathbf{w}_j$ are the item embedding vectors defined in \cref{eq:embed}.

The set-specific contribution to the log-likelihood for using Dynamic DPPs is defined as follows:
\begin{align}
\mathcal{L}_\text{Set}&(\omega, \theta; \hist(T)) := \sum_{i=1}^N \log p_\omega(x_i \sep t_i, \mathbf{h}_\theta(t_i)) \\
& = \sum_{i=1}^N \log \text{det}(\mathbf{L}_{x_i}(t_i)) - \log \text{det}(\mathbf{L}(t_i) + \mathbf{I}). \notag
\end{align}
See \cref{sec:exp_details_dpp} for discussions on parallelizing computation with varying sizes of sets $x_i$.

\paragraph{Optimization Details}
To model the conditional set distribution for each of the Dynamic Bernoulli and DPP models, we learn parameters $\theta, \phi,$ and $\omega$ jointly using stochastic gradient methods. Given a dataset of $M$ sequences over varying timespans, ${\mathcal{D} := \{\hist_i(T_i)\}_{i=1}^M}$, we minimize the negative log-likelihood with the following objective function: $-\mathcal{L}(\theta, \phi, \omega; \mathcal{D}) = -\sum_{i=1}^M \left[\mathcal{L}_\text{Set}(\omega, \theta; \hist_i) + \mathcal{L}_\text{Time}(\phi, \theta; \hist_i)\right].$ 
Should the integral in $\mathcal{L}_\text{Time}$ not be tractable for a given recurrent MTPP base model, we approximate it using Monte-Carlo samples as described in \citet{mei2017neural}.

\section{PROBABILISTIC QUERIES}
By being fully probabilistic over entire sequences of events $\hist(T)$, we arrive at a model that possesses beliefs about future trajectories that encompass more than just the immediate next event's set-value and/or time of occurrence. For instance, MTPPs in general are able to assign probabilities to when the next time a specific mark will occur, often referred to as the \emph{hitting time}, as well as how likely one mark will occur before another mark. Due to the autoregressive nature of MTPPs, this information is not readily available; however, \citet{boyd2023probabilistic} demonstrated how to efficiently estimate these probabilistic beliefs using importance sampling.

In this section, we will demonstrate how to adapt these importance sampling techniques for estimating hitting time and $A$-before-$B$ queries to take advantage of the structure inherent to the set-valued marks. This will enable us to estimate probabilities concerning \emph{item-level queries}. These queries include examples such as ``When will item $k$ occur next?'' and ``Will item $i$ occur before item $j$?'' For brevity, we will discuss only the resulting estimators (for these two kinds of queries) in the context of the Dynamic Bernoulli model for sets.

In the context of prior work on querying with MTPPs, we build on the proposal distribution from \citet{boyd2023probabilistic}, but with several novel aspects. 
(\textit{i}) Since we ask \textit{item-level queries} in the context of \textit{sets}, direct application of \citet{boyd2023probabilistic}'s method yields intractable estimators, for the same reason that directly modeling $\lambda^*_x(t)$ for $2^K$ set-specific intensities is not practical. Thus, we model the structure of sets so that the conditional set distributions provide convenient and tractable marginalization properties for relevant subsets, making query estimation feasible. To achieve this, we discuss two equivalent sampling methods in \cref{sec:query_derivation_sampling}. (\textit{ii}) For $A$-before-$B$ queries, we take the set-based scenario into account where multiple items can occur at the same time: our resulting estimator is unbiased and more accurate (\cref{sec:query_derivation_ab}), unlike \citet{boyd2023probabilistic}'s estimator for single-mark MTPPs.
(\textit{iii}) An additional novelty is the application of the new set-based query estimators to the problem of model selection, which we discuss in more detail in \cref{sec:experiments}.

\subsection{Hitting Time Queries}

\begin{figure}
    \centering
    \includegraphics[width=0.7\columnwidth]{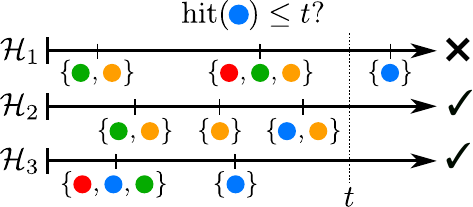}
    \caption{Example sequences for hitting time queries.}
    \label{fig:query_hit}
\end{figure}
Without loss of generality, we suppress the notation for conditioning on partially observed sequences and present all derivations and notation below for unconditional queries.  We also suppress all parameter notation (e.g., replacing $p_\omega$ with $p$), as our focus in querying is on queries with a pre-trained model. Additional details for our proposed importance sampling framework (e.g., full proofs) are provided in \cref{sec:query_derivation}.

We are interested in the distribution of the first time of occurrence for any item in the non-empty subset $A\subseteq\{1, \dots, K\}$ denoted as $\hit(A)$;
see \cref{fig:query_hit} for an illustration. We can define a naive estimator by simply representing any query as an expected value,
\begin{align}
p(\hit(A) \leq t) & = \E_{\hist[0, t]}\left[\ind(\hit(A) \leq t)\right], 
\end{align}
and approximate it using a Monte-Carlo estimate of $\ind(\hit(A) \leq t)$ with $\hist[0, t]$ drawn from the MTPP. The variance of this estimator can be vastly reduced (thus, improving the relative efficiency) by utilizing importance sampling.
Similar to \citet{boyd2023probabilistic}, we define the proposal distribution $q$ as an MTPP with intensity $\mu_x^*(t) = \ind(A \cap x = \emptyset
)\lambda_x^*(t)$. It follows that the total proposal intensity, which is used for sampling \citep{ogata1981lewis}, is $\mu^*(t) = \lambda^*(t)p^*(A^c \sep t)$ where $A^c$ is the complement of $A$ and $p^*(A \sep t)$ is the probability of having any of the items in $A$ occur at time $t$: $p^*(A \cap x \neq \emptyset
\sep t)$. For the Dynamic Bernoulli model, this simplifies to $p^*(A \sep t) = 1 - \prod_{k \in A} (1-p^*(k \sep t))$.

It can then be shown that the CDF of the hitting time takes the following form:
\begin{align}
p(\hit(A) \leq t) & := 1 - \E^q_{\hist[0,t]}\!\left[\exp\!\left(\!-\!\!\int_0^t\!\! \lambda^*(s)p^*(A \sep s) ds\!\right)\!\right] \notag
\end{align}
where $\E^q$ is the expected value with respect to the proposal distribution $q$. This also produces an estimator through a Monte-Carlo estimate using $\hist(t)\sim q$.

\subsection{$A$-before-$B$ queries} \label{sec:ab_query}
Let $A$ and $B$ be two non-overlapping subsets of items. We are interested in which subset has an item that occurs before any items in the other subset occur. This can be analyzed by asking $p(\hit(A) < \hit(B))$. Since we are estimating the probability of this scenario with finite-length sampled sequences, to remain unbiased we must slightly alter the query of interest to ${p(\hit(A) < \hit(B), \hit(A) \leq t)}$.
As before, we can easily derive a naive estimator for this query:
\begin{align}
    p(\hit(A) < &\hit(B), \hit(A) \leq t) \notag \\
    &= \E_{\hist[0, t]}\left[\ind(\hit(A) < \hit(B), \hit(A) \leq t)\right]. \notag
\end{align}
Using a proposal marked intensity $\mu_x^*(t) := \ind((A \cup B) \cap x = \emptyset
)\lambda_x^*(t)$ with total intensity $\mu^*(t) := \lambda^*(t)p^*((A \cup B)^c \sep t)$ allows for the following importance sampled estimator:
\begin{align}
& p(\hit(A) < \hit(B), \hit(A) \leq t) \label{eq:is_ab}\\
& = \E_{\hist[0, t]}^q \left[\int_0^t \exp\left(-\int_0^s \lambda_{A \cup B}^*(s')ds'\right)\lambda_{A\cap\lnot B}^*(s)ds\right], \notag
\end{align}
where the marked intensities use the factorization in \cref{eq:factor} and $p^*(A \cap \lnot B \sep t)$ is the probability of any item in $A$ occurring and no item in $B$ occurring, which is $p^*(A \cap x \neq \emptyset, B \cap x = \emptyset \sep t)$.
It simplifies to ${p^*(A \cap \lnot B \sep t) = p^*(A \sep t)(1-p^*(B \sep t))}$ for the Dynamic Bernoulli model.

There are four different scenarios that can occur when comparing $\hit(A)$ and $\hit(B)$ up to a finite length of time $t$: (\textit{i}) $(\hit(A) < \hit(B), \hit(A) \leq t)$, (\textit{ii}) $(\hit(B) < \hit(A), \hit(B) \leq t)$, (\textit{iii}) $\hit(A) = \hit(B) \leq t$, and (\textit{iv}) $(\hit(A) > t, \hit(B) > t)$. 
Importance sample estimators for (\textit{ii}) and (\textit{iii}) can be achieved by swapping $A \cap \lnot B$ in \cref{eq:is_ab} with $\lnot A \cap B$ and $A \cap B$ respectively. (\textit{iv}) can be estimated by subtracting estimates of (\textit{i-iii}) from 1, due to the law of total probability.

\section{EXPERIMENTAL RESULTS}\label{sec:experiments}

We investigate the prediction performance of our proposed dynamic models and the querying efficiency of our proposed sampling methods using four real-world datasets. Our results show that the proposed dynamic models systematically outperform static alternatives and baselines in terms of the log-likelihood of test sequences. For query-answering, we also
demonstrate that importance sampling is orders of magnitude more efficient in the number of samples (to achieve the same variance) compared to naive sampling, while maintaining (or even reducing) wall-clock runtime.
Furthermore, both test sequence log-likelihood and our proposed query-based log-likelihood favor dynamic models that jointly model time and sets. Our code and data are publicly available\footnote{\href{https://github.com/yuxinc17/set_valued_mtpp}{https://github.com/yuxinc17/set\_valued\_mtpp}}.

\subsection{Datasets}
\cref{tab:datasets} summarizes  the four real-world user behavior datasets
in our experiments, where each sequence represents a user.
\textbf{Instacart}\footnote{\href{https://www.kaggle.com/competitions/instacart-market-basket-analysis}{https://www.kaggle.com/competitions/instacart-market-basket-analysis}} contains samples of grocery orders for customers.
We preprocessed the data into sequences of time-stamped sets of products that belong to 21 departments that we interpret as items. \textbf{Last.fm} \citep{mcfee2012million}
records the listening habits of approximately 1000 users, where the tracks are mapped to 15 genres jointly by artist and title via monthly Discogs Release\footnote{\href{https://discogs.com}{https://discogs.com}}. \textbf{MovieLens 25M} \citep{harper2015movielens} has movie ratings from users and we used the  data from  2016; the 20 different genres of movies are treated as items. For the Last.fm and MovieLens datasets, a track or a movie can have multiple genres. Each set of genres associated with a track or a movie is interpreted as an event. Finally, \textbf{MOOC} \citep{kumar2019predicting} includes 97 distinct user actions involving online course activities that are considered as items. For all datasets, we randomly selected 75\% of sequences for training, 10\% for validation, and 15\% for test. The test split was only used for reporting the results. Additional dataset details are described in \cref{sec:exp_details_data}.

\begin{table}
\centering
\caption{Summary Statistics for Datasets}
\begin{tabular}{l c c c c}
\toprule
Dataset &  \# Seq. &  $K$ & $T_{max}$ & Avg.length  \\
\midrule
Instacart & 174,615 & 21 & 366 & 17 \\
Last.fm & 10,705 & 15 & 744 & 207 \\
MovieLens & 11,198 & 20 & 8,781 & 65  \\
MOOC & 6,892 & 97 & 715 & 52 \\
\bottomrule
\end{tabular}
\label{tab:datasets}
\end{table}

\subsection{Baselines and Model Fitting}
In our experiments, we evaluate the predictions and query computation for models of different complexities, including our proposed dynamic models, simple baselines, and ablations of our models that lie between these endpoints. The models differ in terms of how the temporal and set components in each are modeled. Note that existing MTPP models assume no more than one event of a single type can occur simultaneously, and therefore are not directly comparable because likelihoods would not be commensurate.

Our simplest baseline uses a homogeneous Poisson model as the temporal component and a static Bernoulli model for the set distribution (where the Bernoulli probabilities correspond to the marginal probabilities in the dataset for each item), referred to below as the {\it StaticB-Poisson} model. This simple baseline provides useful context for evaluating the effectiveness of more complex models for set-valued data over time.

In terms of our proposed models, for the temporal component we use the neural Hawkes (NH) model \citep{mei2017neural} as a specific instantiation of the recurrent MTPP component. In the Bernoulli variants of our model, this is coupled with (\textit{i}) our proposed Dynamic Bernoulli model for the set-component or (\textit{ii}) the marginal (static) Bernoulli option as a baseline (same model for sets as the Poisson baseline), referred to below as {\it DynamicB-NH} and {\it StaticB-NH} respectively. In the DPP variants, we couple the NH temporal component of the model with (\textit{i}) the Dynamic-DPP approach for modeling the set component, or (\textit{ii}) again with the marginal (static) DPP, referred to as {\it DynamicDPP-NH} and {\it StaticDPP-NH} respectively.
The static versions of our models
can be viewed as ablations that model time and set structure separately, where the sets are not conditioned on hidden states and therefore invariant of time. More details on models and training procedures are \cref{sec:exp_details_model}.

\subsection{Results: Test Sequence Log-Likelihood}

\cref{tab:log_likelihood} summarizes the average log-likelihood results for all test sequences across the four datasets.
 For all datasets, our dynamic models systematically produce significantly lower negative test log-likelihoods $-\mathcal{L}$ compared to the static baselines. The neural temporal component (NH) is also clearly superior to the Poisson baseline.
 
 In comparing the Dynamic Bernoulli and Dynamic DPP models, the differences in log-likelihoods $\mathcal{L}$ are very small relative to the scale of log-likelihood improvement in going from static to dynamic models. Further, as mentioned earlier, the DPP variant scales much more poorly as a function of the number of items $K$ (and was not scalable in our experiments for the MOOC dataset with $K=97$). Given these observations, in our experiments in the remainder of the paper, we focus on the Dynamic Bernoulli model since it represents a useful practical trade-off between prediction performance and computational cost. Note that our proposed querying scheme, based on importance sampling, can also be used with the Dynamic DPP model, and we provide a sketch of the general approach in \cref{sec:query_derivation_sampling}.
 
 We also report results in \cref{tab:log_likelihood} for the decomposition of the full log-likelihood $\mathcal{L}$ into its components $\mathcal{L}_\text{Time}$ and $\mathcal{L}_\text{Set}$ (\cref{eq:logl}). From these results, we see that static models can have slightly better performance in terms of just the temporal component of the log-likelihood $\mathcal{L}_\text{Time}$. This is because the RNN for static models focuses only on modeling $\mathcal{L}_{\text{Time}}$; marginal set distributions are learned from the data. For dynamic models, sets are conditioned on history; the RNN is used for both time and set prediction. Given that in these experiments the RNN capacity of the dynamic model is the same as the static model, the static models can achieve slightly better results for $\mathcal{L}_{\text{Time}}$; this gap can be reduced by increasing the capacity of the RNN. Empirically, for example,  doubling the hidden state size of the Dynamic Bernoulli model for MovieLens achieves negative test log-likelihoods $-\mathcal{L}_{\text{Time}} = -200.94$ and $-\mathcal{L}_{\text{Set}} = 432.54$. Further discussion is in \cref{sec:exp_results_rnn}.

\begin{table}[h]
\
\caption{Negative test sequence log-likelihood, $-\mathcal{L}$ from \cref{eq:logl},  across four datasets, with different static and dynamic variants of models. Also shown is the decomposition of each  $-\mathcal{L}$ score into time $\mathcal{L}_\text{Time}$ and set $\mathcal{L}_\text{Set}$  components. We highlight the results for $-\mathcal{L}$  to denote the first (bold) and second (underline) best-performing models overall for each dataset. 
``\textbf{--}'' entries indicate that a method required greater memory resources than were available.}

\centering
\resizebox{0.9\columnwidth}{!}{ 
\begin{tabular}{p{3cm} r r r}
\toprule
Dataset\hspace*{\fill}Model &  \multicolumn{1}{c}{$-\mathcal{L} (\downarrow)$} & \multicolumn{1}{c}{$-\mathcal{L}_\text{Time}(\downarrow)$} & \multicolumn{1}{c}{$-\mathcal{L}_\text{Set}(\downarrow)$}\\
\midrule
\multicolumn{1}{l}{\textit{Instacart}} \\
\multicolumn{1}{r}{StaticB-Poisson} & 205.11 & 58.17 & 146.94 \\
\arrayrulecolor{gray}\cmidrule(l{2em}r{0.5em}){1-4}
\multicolumn{1}{r}{StaticB-NH} & 198.22 & 51.30 & 146.92 \\
\multicolumn{1}{r}{StaticDPP-NH} & 203.35 & 51.37 & 151.98 \\
\cmidrule(l{2em}r{0.5em}){1-4}
\multicolumn{1}{r}{DynamicB-NH} & \textbf{168.04} & 51.46 & 116.58 \\
\multicolumn{1}{r}{DynamicDPP-NH} & \underline{170.68} & 51.41 & 119.27 \\
\arrayrulecolor{black}
\midrule
\multicolumn{1}{l}{\textit{Last.fm}} \\
\multicolumn{1}{r}{StaticB-Poisson} & 1027.14 & 377.17 & 649.97 \\
\arrayrulecolor{gray}\cmidrule(l{2em}r{0.5em}){1-4}
\multicolumn{1}{r}{StaticB-NH} & 415.00 & -234.91 & 649.92 \\
\multicolumn{1}{r}{StaticDPP-NH} & 411.68 & -235.70 & 647.38 \\
\cmidrule(l{2em}r{0.5em}){1-4}
\multicolumn{1}{r}{DynamicB-NH} & \underline{259.08} & -223.59 & 482.67 \\
\multicolumn{1}{r}{DynamicDPP-NH} & \textbf{258.82} & -223.33 & 482.15 \\
\arrayrulecolor{black}

\midrule
\multicolumn{1}{l}{\textit{MovieLens}} \\
\multicolumn{1}{r}{StaticB-Poisson} & 741.55 & 276.49 & 465.07 \\
\arrayrulecolor{gray}\cmidrule(l{2em}r{0.5em}){1-4}
\multicolumn{1}{r}{StaticB-NH} & 263.78  & -201.19 & 464.97 \\
\multicolumn{1}{r}{StaticDPP-NH} & 259.95 & -203.95 & 463.90 \\
\cmidrule(l{2em}r{0.5em}){1-4}
\multicolumn{1}{r}{DynamicB-NH} & \underline{236.80} & -195.78 & 432.58 \\
\multicolumn{1}{r}{DynamicDPP-NH} & \textbf{236.35} & -194.15 & 430.50 \\
\arrayrulecolor{black}

\midrule
\multicolumn{1}{l}{\textit{MOOC}}  \\
\multicolumn{1}{r}{StaticB-Poisson} & 439.74 & 169.77 & 269.97 \\
\arrayrulecolor{gray}\cmidrule(l{2em}r{0.5em}){1-4}
\multicolumn{1}{r}{StaticB-NH} & \underline{189.54} & -81.66 & 271.20 \\
\multicolumn{1}{r}{StaticDPP-NH} & \multicolumn{1}{c}{\textbf{--}} & \multicolumn{1}{c}{\textbf{--}} & \multicolumn{1}{c}{\textbf{--}} \\
\cmidrule(l{2em}r{0.5em}){1-4}
\multicolumn{1}{r}{DynamicB-NH} & \textbf{45.30} & -77.06 & 122.35 \\
\multicolumn{1}{r}{DynamicDPP-NH} & \multicolumn{1}{c}{\textbf{--}} & \multicolumn{1}{c}{\textbf{--}} & \multicolumn{1}{c}{\textbf{--}} \\
\arrayrulecolor{black}

\bottomrule
\end{tabular}
}
\label{tab:log_likelihood}
\end{table}

\subsection{Results: Querying}\label{sec:main_results_querying}

We evaluate our querying methods from two perspectives: (\textit{i}) the relative efficiency of the importance estimate compared to naive sampling, 
and (\textit{ii}) the average query log-likelihood with respect to a trained model.

\begin{figure}
    \centering
    \includegraphics[trim={0 0.2cm 0 0}, width=0.9\columnwidth]{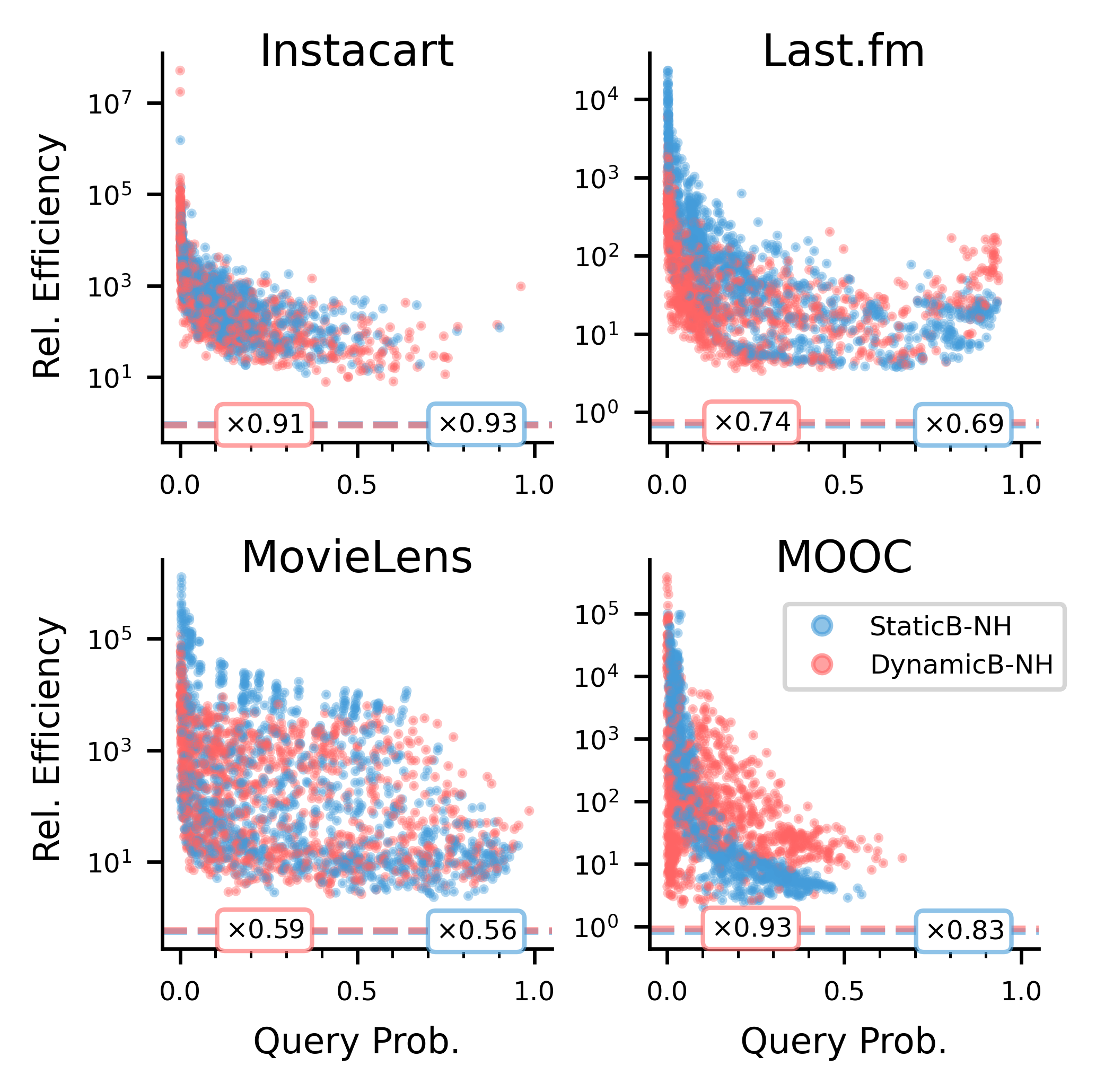}
    \caption{Relative efficiency for queries of the form $p(\hit(A) \leq t \mid \hist)$ for two model variants. Blue and red dashed lines refer to the multiplicative runtime of importance sampling compared to naive sampling.}
    \label{fig:hit_eff}
\end{figure}

\paragraph{Efficiency for Hitting Time Queries}

Relative efficiency is defined as the ratio between the variance of naive estimates and the variance of importance estimates, and is widely used in assessing two \textit{unbiased} estimators \citep{van2000asymptotic}. This number can be interpreted as the number of samples that would be required for naive samples to achieve the same level of accuracy in estimation, e.g., to a particular degree of numerical precision in absolute error for the  estimate.
 
For hitting time queries we are interested in the probability $p(\hit(A)\leq t \mid \hist)$, defined as the probability that one or more items in the set $A$ are observed before a fixed time $t$ (where $A$ and $t$ are hyperparameters of the query), conditioned on sequence history. While our importance sampling approach is compatible with subsets of any size, in the experiments below we report on the simplest case where $A$ contains a single item. %
We also experiment with varying numbers of items in $A$ and analyze the runtime of importance sampling and naive approach, and find that importance sampling takes less wall-clock time per sample than naive sampling with a gain in time for an increasing number of items in $A$; see \cref{sec:exp_results_runtime} for more results and discussion.

We use 1000 queries, 1000 importance samples, and 2000 integration points in our experiments for all datasets and models. For each test sequence, we define a query $\prob (\hit(A) \leq t | \hist)$ by conditioning on the first five events $\hist$ and $t = \min((\tau_6 - \tau_5) \times 10, 10)$. The single item in $A$ is chosen from existing items in $\hist$ for the MOOC dataset due to the large number of items, and is chosen randomly from all possible items for the other three datasets. Note that the item is not guaranteed to occur in the remaining observed sequence. 

\cref{fig:hit_eff} plots the relative efficiency for both static and dynamic models on four datasets. We observe that importance sampling is vastly more sample-efficient ($y$-axis) for both static and dynamic models, where the gains have some dependence on the query probability being estimated ($x$-axis). In other words, importance samples significantly reduce variance compared to naive samples, while maintaining comparable runtimes in terms of wall-clock time (dashed lines).

\paragraph{Efficiency for $A$-before-$B$ Queries}

Given the history of a sequence, we are interested in the probability of the order of the first occurrence of two items $A, B$ during an observation window into the future.
We define the  $A$-before-$B$ query as $p(\hit(A) < \hit(B), \hit(A) \leq t \mid \hist)$.
We adopt the same general settings as described above for hitting time queries with
one difference: we choose two random items $A$ and $B$ from the existing items in $\hist$ for all datasets except for the Instacart dataset, and choose the two items randomly from all items for the Instacart datasets.

\begin{figure}
    \centering
    \includegraphics[trim={0 0.2cm 0 0}, width=0.9\columnwidth]{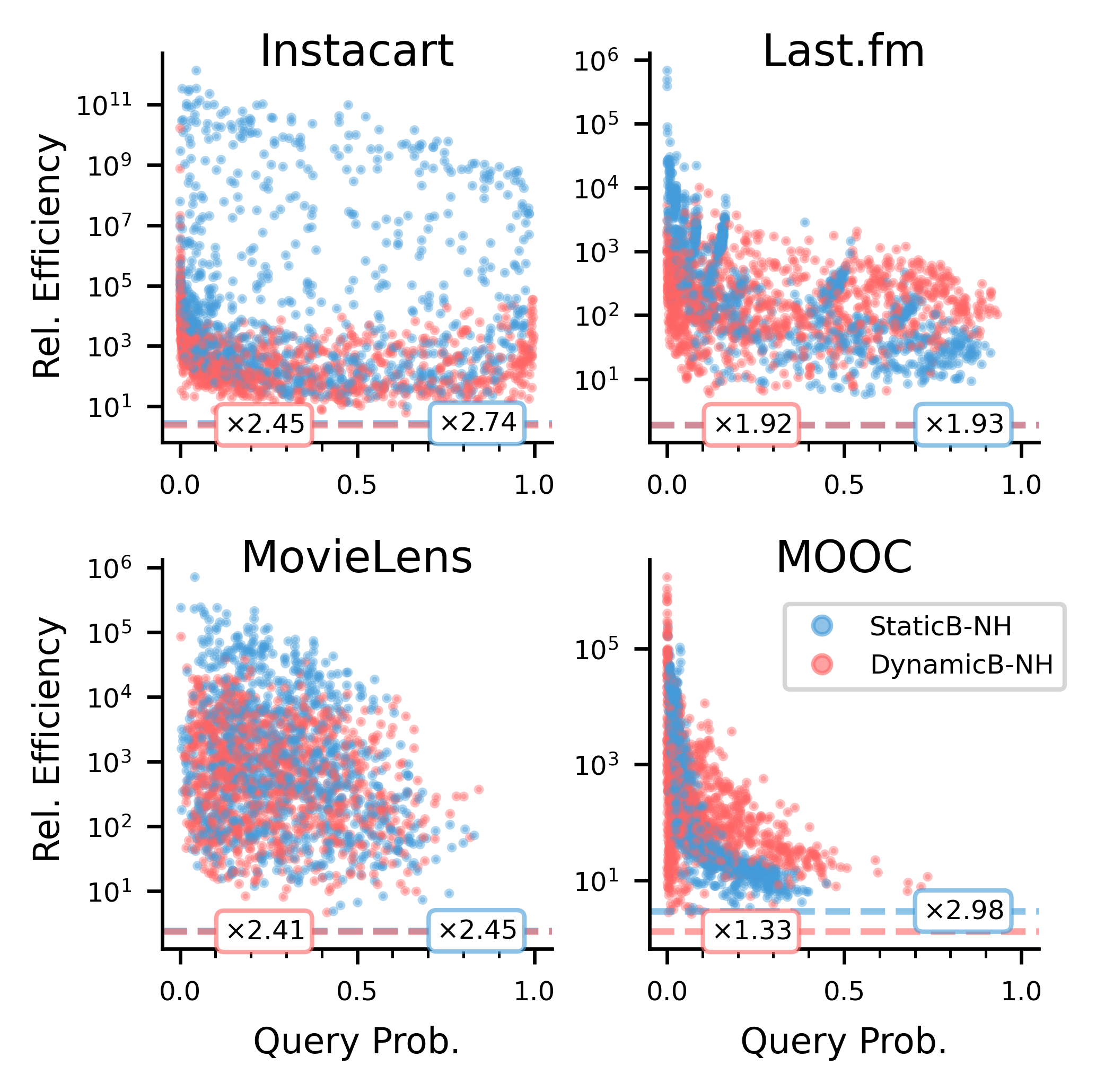}
    \caption{Relative efficiency results for the query $p(\hit(A) < \hit(B), \hit(A) \leq t \mid \hist)$ with the same format as \cref{fig:hit_eff}.}
    \label{fig:ab_eff}
\end{figure}

Relative efficiency results for $A$-before-$B$ queries are shown in \cref{fig:ab_eff}. As with hitting time queries we see orders of magnitude improvements in efficiency using importance sampling, again with comparable wall-clock computation times per sample. 

Note that the relative runtime between naive and importance samples varies across queries and datasets. Importance samples can be sometimes faster (0.5$\times$ to 0.9$\times$ in \cref{fig:hit_eff}), but this is not always the case (2$\times$ to 3$\times$, i.e., slower, in \cref{fig:ab_eff}). (\textit{i}) Importance samples can in some situations be {\it faster} because the proposal distribution essentially zeros out some intensities. Consequently, fewer events are sampled on average. (\textit{ii}) Importance samples can also sometimes be {\it slower} than naive sampling, because additional computations such as integral estimates are performed for each sample to estimate query probability. There is a trade-off between sampling fewer events and more computation per sample. Additional discussion related to runtimes can be found in \cref{sec:exp_results_runtime}. 

In general, we have observed empirically that importance samples consistently have runtimes that are comparable to those of naive sampling. This is true for both the NH model discussed above as well as for other instantiations of recurrent MTPP models such as RMTPP (\cref{fig:eff_rmtpp} in \cref{sec:exp_result_rmtpp}) and slightly different model architectures (\cref{fig:hit_eff_2layers,fig:ab_eff_2layers}) in \cref{sec:exp_results_ablation}.

\begin{figure}
    \centering
        \includegraphics[trim={0.1cm 0.15cm 0 0},clip,width=0.9\columnwidth]{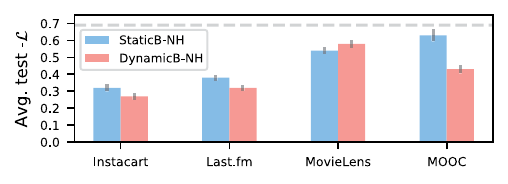}
    \includegraphics[trim={0.1cm 0.1cm 0 0}, clip, width=0.9\columnwidth]{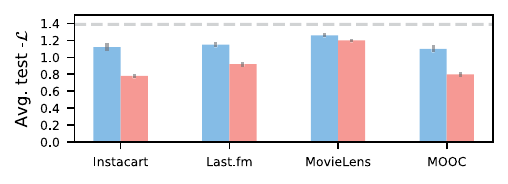}
    \caption{Average negative test log-likelihood ($\pm 1$ std. dev.) of hitting time (top) 
    and  $A$-before-$B$ (bottom) queries across 4 datasets.
    Lower values are better: the lower bound is 0 and the upper dashed line is the negative log-likelihood of randomly guessing outcomes.}
    \label{fig:query_ll}
\end{figure}

\paragraph{Model Comparison using Query Likelihoods}
Model comparisons for probabilistic sequential models are often made by computing the likelihood of test sequences, involving sums of log-probabilities for observed ``next events'' conditioned on observed histories (as in \cref{tab:log_likelihood}). Our querying approach allows us in principle to compute log-likelihoods of propositions that are beyond next-event prediction, e.g., how well do two models compare in terms of their log-likelihoods for predicting actual hitting times on a particular dataset?

More specifically, for hitting times we can define a log-likelihood score for $p(\hit(A)\leq t \mid \hist)$  as $\mathcal{L} = y\hat{p} + (1-y)(1-\hat{p})$,
where $\hat{p}$ is an importance sampling estimate and $y:=\ind{(\hit(A)\leq t \mid \hist)}$.
We compute a similar log-likelihood for $A$-before-$B$ queries, but we account for the four different outcomes as outlined in \cref{sec:ab_query}.

\cref{fig:query_ll} shows the results for computing log-likelihood scores (averaged over the same queries described earlier) for both hitting time probabilities (top) and $A$-before-$B$ probabilities (bottom). Each bar-plot compares the StaticB-NH model with the DynamicB-NH model across the 4 datasets used in our earlier experiments.
The dynamic model is strongly favored over the static model on three of the datasets and the difference is not statistically significant for MovieLens  (overlapping standard deviations), 
in broad agreement with the results in \cref{tab:log_likelihood}.
Note that this type of model comparison is practical computationally using importance sampling, but not with naive sampling, 
opening up the option of richer data-driven model comparisons for MTPPs. %

\section{CONCLUSION}
In this work we proposed a framework that models the joint likelihood of continuous-time set sequences for both temporal and set distributions, built on recurrent MTPP models. Across four real-world user behavior datasets, our proposed model achieves better predictive performance than baselines, both over sequences and various probabilistic queries. In addition, our importance sampling approach is orders of magnitude more efficient than naive methods in answering probabilistic queries in the context of subsets. 
Future work could include exploration of more complex set structures (e.g., conditionally modeling positive and negative item-inclusion correlations) and extensions to modeling subsets of continuous values.

\subsection*{ACKNOWLEDGEMENTS}
We thank the reviewers for their useful suggestions for improving the paper.
This work was supported by National Science Foundation Graduate
Research Fellowship grant DGE-1839285 (AB), by the National Science Foundation under award number 1900644 (PS),
by the National Institute of Health under awards R01-AG065330-02S1 and R01-LM013344 (YC and PS), by the HPI Research Center in Machine Learning and Data Science at UC Irvine (YC and PS), and by Qualcomm Faculty awards (PS).

\bibliography{references}

\begin{thebibliography}{}

\bibitem[Bacry and Muzy, 2014]{bacry2014hawkes}
Bacry, E. and Muzy, J.-F. (2014).
\newblock Hawkes model for price and trades high-frequency dynamics.
\newblock {\em Quantitative Finance}, 14(7):1147--1166.

\bibitem[Bonnet et~al., 2023]{bonnet2023inference}
Bonnet, A., Martinez~Herrera, M., and Sangnier, M. (2023).
\newblock Inference of multivariate exponential {H}awkes processes with inhibition and application to neuronal activity.
\newblock {\em Statistics and Computing}, 33(4):91.

\bibitem[Boyd et~al., 2023]{boyd2023probabilistic}
Boyd, A., Chang, Y., Mandt, S., and Smyth, P. (2023).
\newblock Probabilistic querying of continuous-time event sequences.
\newblock In {\em International Conference on Artificial Intelligence and Statistics}, pages 10235--10251. PMLR.

\bibitem[Brillinger, 1975]{brillinger1975identification}
Brillinger, D.~R. (1975).
\newblock The identification of point process systems.
\newblock {\em The Annals of Probability}, pages 909--924.

\bibitem[Daley and Vere-Jones, 2003]{daley2003introduction}
Daley, D.~J. and Vere-Jones, D. (2003).
\newblock {\em An Introduction to the Theory of Point Processes: Volume I: Elementary Theory and Methods}, page 274.
\newblock Springer.

\bibitem[Deshpande et~al., 2021]{deshpande2021long}
Deshpande, P., Marathe, K., De, A., and Sarawagi, S. (2021).
\newblock Long horizon forecasting with temporal point processes.
\newblock In {\em Proceedings of the 14th ACM International Conference on Web Search and Data Mining}, pages 571--579.

\bibitem[Du et~al., 2016]{du2016recurrent}
Du, N., Dai, H., Trivedi, R., Upadhyay, U., Gomez-Rodriguez, M., and Song, L. (2016).
\newblock Recurrent marked temporal point processes: Embedding event history to vector.
\newblock In {\em Proceedings of the 22nd ACM SIGKDD International Conference on Knowledge Discovery \& Data Mining}, pages 1555--1564.

\bibitem[Giudici et~al., 2023]{giudici2023network}
Giudici, P., Pagnottoni, P., and Spelta, A. (2023).
\newblock Network self-exciting point processes to measure health impacts of {COVID}-19.
\newblock {\em Journal of the Royal Statistical Society Series A: Statistics in Society}, 186(3):401--421.

\bibitem[Gupta et~al., 2021]{gupta2021learning}
Gupta, V., Bedathur, S., Bhattacharya, S., and De, A. (2021).
\newblock Learning temporal point processes with intermittent observations.
\newblock In {\em International Conference on Artificial Intelligence and Statistics}, pages 3790--3798. PMLR.

\bibitem[Harper and Konstan, 2015]{harper2015movielens}
Harper, F.~M. and Konstan, J.~A. (2015).
\newblock The {MovieLens} datasets: History and context.
\newblock {\em ACM Transactions on Interactive Intelligent Systems (TIIS)}, 5(4):1--19.

\bibitem[Hawkes, 1971]{hawkes1971spectra}
Hawkes, A.~G. (1971).
\newblock Spectra of some self-exciting and mutually exciting point processes.
\newblock {\em Biometrika}, 58(1):83--90.

\bibitem[Hawkes, 2018]{hawkes2018hawkes}
Hawkes, A.~G. (2018).
\newblock Hawkes processes and their applications to finance: a review.
\newblock {\em Quantitative Finance}, 18(2):193--198.

\bibitem[Holbrook et~al., 2022]{holbrook2022viral}
Holbrook, A.~J., Ji, X., and Suchard, M.~A. (2022).
\newblock From viral evolution to spatial contagion: a biologically modulated {H}awkes model.
\newblock {\em Bioinformatics}, 38(7):1846--1856.

\bibitem[Hu and He, 2019]{hu2019sets2sets}
Hu, H. and He, X. (2019).
\newblock Sets2sets: Learning from sequential sets with neural networks.
\newblock In {\em Proceedings of the 25th ACM SIGKDD International Conference on Knowledge Discovery \& Data Mining}, pages 1491--1499.

\bibitem[Kingma and Ba, 2015]{DBLP:journals/corr/KingmaB14}
Kingma, D.~P. and Ba, J. (2015).
\newblock Adam: A method for stochastic optimization.
\newblock {\em International Conference on Learning Representations (ICLR)}.

\bibitem[Kulesza and Taskar, 2010]{kulesza2010structured}
Kulesza, A. and Taskar, B. (2010).
\newblock Structured determinantal point processes.
\newblock {\em Advances in Neural Information Processing Systems}, 23:1171--1179.

\bibitem[Kulesza et~al., 2012]{kulesza2012determinantal}
Kulesza, A., Taskar, B., et~al. (2012).
\newblock Determinantal point processes for machine learning.
\newblock {\em Foundations and Trends{\textregistered} in Machine Learning}, 5(2--3):123--286.

\bibitem[Kumar et~al., 2019]{kumar2019predicting}
Kumar, S., Zhang, X., and Leskovec, J. (2019).
\newblock Predicting dynamic embedding trajectory in temporal interaction networks.
\newblock In {\em Proceedings of the 25th ACM SIGKDD International Conference on Knowledge Discovery \& Data Mining}, pages 1269--1278.

\bibitem[Lee et~al., 2022]{sarita2022}
Lee, S.~D., Shen, A.~A., Park, J., Harrigan, R.~J., Hoff, N.~A., Rimoin, A.~W., and Paik~Schoenberg, F. (2022).
\newblock Comparison of prospective {H}awkes and recursive point process models for {E}bola in {DRC}.
\newblock {\em Journal of Forecasting}, 41(1):201--210.

\bibitem[Ma et~al., 2021]{ma2021bridging}
Ma, Y., Liu, G., and Deoras, A. (2021).
\newblock Bridging recommendation and marketing via recurrent intensity modeling.
\newblock In {\em International Conference on Learning Representations}.

\bibitem[Macchi, 1975]{macchi1975coincidence}
Macchi, O. (1975).
\newblock The coincidence approach to stochastic point processes.
\newblock {\em Advances in Applied Probability}, 7(1):83--122.

\bibitem[McFee et~al., 2012]{mcfee2012million}
McFee, B., Bertin-Mahieux, T., Ellis, D.~P., and Lanckriet, G.~R. (2012).
\newblock The million song dataset challenge.
\newblock In {\em Proceedings of the 21st International Conference on World Wide Web}, pages 909--916.

\bibitem[Mei and Eisner, 2017]{mei2017neural}
Mei, H. and Eisner, J.~M. (2017).
\newblock The neural {H}awkes process: A neurally self-modulating multivariate point process.
\newblock {\em Advances in Neural Information Processing Systems}, 30:6757--6767.

\bibitem[Mei et~al., 2019]{mei2019imputing}
Mei, H., Qin, G., and Eisner, J. (2019).
\newblock Imputing missing events in continuous-time event streams.
\newblock In {\em International Conference on Machine Learning}, pages 4475--4485. PMLR.

\bibitem[Ogata, 1981]{ogata1981lewis}
Ogata, Y. (1981).
\newblock On {L}ewis' simulation method for point processes.
\newblock {\em IEEE Transactions on Information Theory}, 27(1):23--31.

\bibitem[Pfaffelhuber et~al., 2022]{pfaffelhuber2022mean}
Pfaffelhuber, P., Rotter, S., and Stiefel, J. (2022).
\newblock Mean-field limits for non-linear {H}awkes processes with excitation and inhibition.
\newblock {\em Stochastic Processes and their Applications}, 153:57--78.

\bibitem[Rambhatla et~al., 2022]{rambhatla2022toward}
Rambhatla, S., Zeighami, S., Shahabi, K., Shahabi, C., and Liu, Y. (2022).
\newblock Toward accurate spatiotemporal {COVID}-19 risk scores using high-resolution real-world mobility data.
\newblock {\em ACM Transactions on Spatial Algorithms and Systems (TSAS)}, 8(2):1--30.

\bibitem[Rendle et~al., 2010]{rendle2010factorizing}
Rendle, S., Freudenthaler, C., and Schmidt-Thieme, L. (2010).
\newblock Factorizing personalized {M}arkov chains for next-basket recommendation.
\newblock In {\em Proceedings of the 19th International Conference on World Wide Web}, pages 811--820.

\bibitem[Shchur et~al., 2019]{shchur2019intensity}
Shchur, O., Bilo{\v{s}}, M., and G{\"u}nnemann, S. (2019).
\newblock Intensity-free learning of temporal point processes.
\newblock In {\em International Conference on Learning Representations}.

\bibitem[Shchur et~al., 2021]{shchur2021neural}
Shchur, O., T{\"u}rkmen, A.~C., Januschowski, T., and G{\"u}nnemann, S. (2021).
\newblock Neural temporal point processes: A review.
\newblock {\em arXiv preprint arXiv:2104.03528}.

\bibitem[Shelton et~al., 2018]{shelton2018hawkes}
Shelton, C., Qin, Z., and Shetty, C. (2018).
\newblock Hawkes process inference with missing data.
\newblock In {\em Proceedings of the AAAI Conference on Artificial Intelligence}, volume~32, pages 6425--6432.

\bibitem[Shou et~al., 2023]{shou2023concurrent}
Shou, X., Gao, T., Subramanian, D., Bhattacharjya, D., and Bennett, K.~P. (2023).
\newblock Concurrent multi-label prediction in event streams.
\newblock In {\em Proceedings of the AAAI Conference on Artificial Intelligence}, volume~37, pages 9820--9828.

\bibitem[T{\"u}rkmen et~al., 2020]{turkmen2020fastpoint}
T{\"u}rkmen, A.~C., Wang, Y., and Smola, A.~J. (2020).
\newblock Fastpoint: Scalable deep point processes.
\newblock In {\em Proceedings of the European Conference on Machine Learning and Knowledge Discovery in Databases: ECML/PKDD}, pages 465--480. Springer.

\bibitem[Van~der Vaart, 2000]{van2000asymptotic}
Van~der Vaart, A.~W. (2000).
\newblock {\em Asymptotic Statistics}, volume~3.
\newblock Cambridge University Press.

\bibitem[Wang et~al., 2008]{wang2008continuous}
Wang, C., Blei, D., and Heckerman, D. (2008).
\newblock Continuous time dynamic topic models.
\newblock In {\em Proceedings of the Twenty-Fourth Conference on Uncertainty in Artificial Intelligence}, pages 579--586.

\bibitem[Wang and McCallum, 2006]{wang2006topics}
Wang, X. and McCallum, A. (2006).
\newblock Topics over time: a non-{M}arkov continuous-time model of topical trends.
\newblock In {\em Proceedings of the 12th ACM SIGKDD international Conference on Knowledge Discovery and Data Mining}, pages 424--433.

\bibitem[Wehrli and Sornette, 2022]{wehrli2022classification}
Wehrli, A. and Sornette, D. (2022).
\newblock Classification of flash crashes using the {H}awkes (p, q) framework.
\newblock {\em Quantitative Finance}, 22(2):213--240.

\bibitem[Xie et~al., 2017]{xie2017deep}
Xie, P., Salakhutdinov, R., Mou, L., and Xing, E.~P. (2017).
\newblock Deep determinantal point process for large-scale multi-label classification.
\newblock In {\em Proceedings of the IEEE International Conference on Computer Vision}, pages 473--482.

\bibitem[Xue et~al., 2022]{xue2022hypro}
Xue, S., Shi, X., Zhang, J., and Mei, H. (2022).
\newblock {HYPRO}: A hybridly normalized probabilistic model for long-horizon prediction of event sequences.
\newblock {\em Advances in Neural Information Processing Systems}, 35:34641--34650.

\bibitem[Yu et~al., 2016]{yu2016dynamic}
Yu, F., Liu, Q., Wu, S., Wang, L., and Tan, T. (2016).
\newblock A dynamic recurrent model for next basket recommendation.
\newblock In {\em Proceedings of the 39th International ACM SIGIR Conference on Research and Development in Information Retrieval}, pages 729--732.

\end{thebibliography}

\clearpage

\section*{Checklist}

 \begin{enumerate}

 \item For all models and algorithms presented, check if you include:
 \begin{enumerate}
   \item A clear description of the mathematical setting, assumptions, algorithm, and/or model. [Yes]
   \item An analysis of the properties and complexity (time, space, sample size) of any algorithm. [Yes]
   \item (Optional) Source code, with specification of all dependencies, including external libraries. [Yes]
 \end{enumerate}

 \item For any theoretical claim, check if you include:
 \begin{enumerate}
   \item Statements of the full set of assumptions of all theoretical results. [Yes]
   \item Complete proofs of all theoretical results. [Not Applicable]
   \item Clear explanations of any assumptions. [Yes]     
 \end{enumerate}

 \item For all figures and tables that present empirical results, check if you include:
 \begin{enumerate}
   \item The code, data, and instructions needed to reproduce the main experimental results (either in the supplemental material or as a URL). [Yes]
   \item All the training details (e.g., data splits, hyperparameters, how they were chosen). [Yes]
         \item A clear definition of the specific measure or statistics and error bars (e.g., with respect to the random seed after running experiments multiple times). [Yes]
         \item A description of the computing infrastructure used. (e.g., type of GPUs, internal cluster, or cloud provider). [Yes]
 \end{enumerate}

 \item If you are using existing assets (e.g., code, data, models) or curating/releasing new assets, check if you include:
 \begin{enumerate}
   \item Citations of the creator If your work uses existing assets. [Yes]
   \item The license information of the assets, if applicable. [Not Applicable]
   \item New assets either in the supplemental material or as a URL, if applicable. [Not Applicable]
   \item Information about consent from data providers/curators. [Not Applicable]
   \item Discussion of sensible content if applicable, e.g., personally identifiable information or offensive content. [Not Applicable]
 \end{enumerate}

 \item If you used crowdsourcing or conducted research with human subjects, check if you include:
 \begin{enumerate}
   \item The full text of instructions given to participants and screenshots. [Not Applicable]
   \item Descriptions of potential participant risks, with links to Institutional Review Board (IRB) approvals if applicable. [Not Applicable]
   \item The estimated hourly wage paid to participants and the total amount spent on participant compensation. [Not Applicable]
 \end{enumerate}

 \end{enumerate}

\onecolumn

\appendix

\section{QUERY DERIVATION}\label{sec:query_derivation}
We derive importance sampling expressions for both hitting time queries and $A$-before-$B$ queries. Our derivations are formulated in the context of subsets, where we focus on relatively simple types of subsets. It is straightforward to extend to more general queries that can be represented as Boolean expressions, as discussed for the case of hitting time queries in \cref{sec:query_derivation_hit}.

\subsection{Hitting Time Queries}\label{sec:query_derivation_hit}
Define the proposal distribution $q$ as an MTPP with intensity $\mu_x^*(t) = \ind(A \cap x = \emptyset)\lambda_x^*(t)$. Hitting time queries can then be derived as follows:

\begin{align*}
p\left( \hit(A) \leq t \right)
&= 1 - p\left( \hit(A) > t \right) \\
&= 1 - p\left( \forall_{(\tau, \set) \in \hist[0,t]} A \notin \set \right)\\
&= 1-\E_{\hist[0,t] \sim q}\left[ \ind\left( \forall_{(\tau, \set) \in \hist[0,t]} A \notin \set \right) \frac{p(\hist[0,t])}{q(\hist[0,t])} \right]\\
&= 1-\E_{\hist[0,t] \sim q}\left[ \frac{p(\hist[0,t])}{q(\hist[0,t])} \right]\\
&= 1-\E_{\hist[0,t] \sim q}\left[ \frac{\prod_{i=1}^\hist \lambda^*_{x_i}(t_i) \exp\left( -\int_0^t \lambda^*(s) ds \right)}{\prod_{i=1}^\hist \lambda^*_{x_i}(t_i) \exp\left( -\int_0^t \int_{x \in \vocab \setminus A} \lambda_x^*(s) dx ds \right)} \right]\\
&= 1-\E_{\hist[0,t] \sim q}\left[ \frac{ \exp\left( -\int_0^t \lambda^*(s) ds \right)}{\exp\left( -\int_0^t \sum_{x \in \vocab \setminus A} \lambda^*_x(s) ds \right)} \right]\\
&= 1-\E_{\hist[0,t] \sim q}\left[ \exp\left( -\int_0^t \sum_{x \in A} \lambda^*_x(s) ds \right) \right] \\
&= 1-\E_{\hist[0,t] \sim q}\left[ \exp\left( 
-\int_0^t \lambda^*(s) \hat{p}_{A}(s)ds \right)\right],
\end{align*}
where $p(\hist)$ denotes the distribution under the original MTPP and $\hat{p}_{A}$ is the probability that any item in $A$ occurs. The hitting time query can be extended to a general DNF representation by deriving the query in terms of subsets when making restrictions on items. For example, $p(\hit(A \lor B) \leq t)$ can be derived by replacing $\hat{p}_{A}$ by $\hat{p}_{A \cup B}$ in the final line of derivation. Additionally, as an example of handling more complex Boolean queries, a disjunction such as $p(\hit((A \wedge B) \lor C) \leq t)$ (for subsets $A, B, C$) can be decomposed using $\cup$, and the conjunction $p(\hit(A \wedge B) \leq t)$ can be calculated as $1-p(\hit(A)>t) - p(\hit(B)>t) + p(\hit(A \cup B)>t)$, as long as the probabilities of these collections of subsets are available.

\subsection{$A$-before-$B$ Queries}\label{sec:query_derivation_ab}
Using a similar proposal distribution $q$ as an MTPP with intensity $\mu_x^*(t) := \ind((A \cup B) \cap x = \emptyset)\lambda_x^*(t)$, the total intensity is $\mu^*(t) := \lambda^*(t)p^*((A \cup B)^c \sep t)$, where $(A \cup B)^c$ is the complement of $(A \cup B)$. 
For $\prob \left( \hit(A) \leq \hit(B), \hit(A) \leq t \right)$, considering the fact that the probability of $\hit(A)=t$ is infinitesimal, then for any fixed $t$ we have:

\begin{align*}
    \prob \left( \hit(A) < \hit(B), \hit(A) \leq t \right) &= \int_0^t  \prob \left( \hit(A) < \hit(B), \hit(A) \in [s, s+ds) \right)  ds \\
    &= \int_0^t \E_{\hist[0, t] \sim p} \left[ \ind\left(\hit(A) < \hit(B)\right) \ind\left( \hit(A) \in [s, s+ds) \right)  \right]ds \\
    &=\int_0^t \E_{\hist[0, t] \sim p}  \left[ \E_{\hist[0,t] \sim p \mid \hist(t)} \left[ \ind\left(\hit(A) < \hit(B)\right) \ind\left( \hit(A) \in [s, s+ds) \right)  \right] \right] ds \\
    &= \int_0^t \E_{\hist[0, t] \sim p}  \left[ \ind\left( \hit(A) \geq t, \hit(B) \geq t \right) \lambda^*_{A \cap (\neg B)} (s)\right] ds \\
    &= \int_0^t \E_{\hist[0, t] \sim q}  \left[ \exp\left( -\int_0^s \lambda^*_{A \cup B} (s') ds'\right)  \lambda^*_{A \cap (\neg B)} (s) \right] ds \\
    &= \E_{\hist[0, t] \sim q}  \left[  \int_0^t \exp\left( -\int_0^s \lambda^*_{A \cup B} (s') ds'\right)  \lambda^*_{A \cap (\neg B)} (s) ds \right].
\end{align*}

The last line holds by applying Fubini’s theorem where the sums are special cases of integrals for discrete measures. Similarly, we can derive:
\begin{align*}
    \prob \left( \hit(A) = \hit(B), \hit(A) \leq t \right) &= \E_{\hist[0, t] \sim q}  \left[  \int_0^t \exp\left( -\int_0^s \lambda^*_{A \cup B} (s') ds'\right)  \lambda^*_{A \cap B} (s) ds \right] \\
    \prob \left( \hit(A) \leq \hit(B), \hit(A) \leq t \right) &= \E_{\hist[0, t] \sim q}  \left[  \int_0^t \exp\left( -\int_0^s \lambda^*_{A \cup B} (s') ds'\right)  \lambda^*_{A} (s) ds \right] \\
    \prob \left( \hit(A) > \hit(B), \hit(B) \leq t \right) &= \E_{\hist[0, t] \sim q}  \left[  \int_0^t \exp\left( -\int_0^s \lambda^*_{A \cup B} (s') ds'\right)  \lambda^*_{(\neg A) \cap B} (s) ds \right],
\end{align*}
and $p(\hit(A) > t, \hit(B) > t)$ can be calculated as the complement of the other three scenarios at any time $t$.

\subsection{Sampling Details}\label{sec:query_derivation_sampling}
We utilize the property $\lambda_x^*(t) := \lambda^*(t)p^*(x \sep t)$ to sample from the proposal distribution. More specifically, for any subset $A$, $\lambda_A^*(t) := \lambda^*(t)p^*(A \sep t)$ where $p^*(A \sep t)$ is determined by the modeling assumption on the set distribution. This can be calculated in directly for Dynamic DPP models as $p^*(A \sep t) = \frac{\text{det}(\mathbf{L}_A(t))}{\text{det}(\mathbf{L}(t) + \mathbf{I})}$ where $\mathbf{L}$ is the $L$-ensemble indexed by the elements of $\vocab$. For Dynamic Bernoulli models, this expression simplifies to $p^*(A \sep t) = 1 - \prod_{k \in A} (1-p^*(k \sep t))$.

There are two equivalent methods to sample from $\lambda_A^*(t)$. The first approach involves a two-step rejection sampling procedure. Similar to sampling from MTPPs where events can only contain one item, we first decide whether to accept a proposal time based on the total intensity $\lambda^*(t)$, and then sample from the set distribution. If the set contains items that are not allowed by the proposal distribution, we reject the event entirely and move forward. Alternatively, a second approach is to reweigh the total intensity if $p^*(A \sep t)$ is in closed form or is simple to compute. Then we can directly sample the proposal time from $\lambda_A^*(t) = \lambda^*(t)p^*(A \sep t)$, and simply re-sample the set if it contains items that are not allowed in the proposal distribution. 

The second approach is more efficient with the models that we propose and is what we used in our experiments in the paper. The first approach may, however, be preferred for more complex queries in situations where the distribution on sets is more complicated, i.e., when $p^*(A \sep t)$ is computationally expensive to estimate.

\section{EXPERIMENTAL DETAILS}\label{sec:exp_details}
All experiments are performed on NVIDIA GeForce RTX 2080 Ti.
\subsection{Data Preprocessing}\label{sec:exp_details_data}
For all datasets, we randomly selected sequences into 75\%/10\%/25\% partitions for training/validation/test. Anonymized user IDs are used for each user.  Event times are in hours (e.g., 1.324 hours) for all datasets except for Instacart where event times are in days (e.g., 2.476 days). Each sequence in each dataset is standardized to start at time $t=0$.

\textbf{Instacart}\footnote{\href{https://www.kaggle.com/competitions/instacart-market-basket-analysis}{https://www.kaggle.com/competitions/instacart-market-basket-analysis}} records customer orders with the order number and the time gap from the last order from a customer. We took the ``prior'' and ``train'' parts of the original dataset. 
This leads to a total of 3,346,083 orders with sets of products from 206,209 users. We generated one sequence per customer and used the union of subsets if a user has multiple orders occurring at the same time. We mapped the products into 21 distinct department IDs, leaving 3,325,578 orders. Finally, we filtered customer sequences to retain sequences that have at least 5 events but no more than 200 events. The resulting dataset contains 174,615 sequences from different customers.  

\textbf{Last.fm}\footnote{\href{http://ocelma.net/MusicRecommendationDataset/lastfm-1K.html}{http://ocelma.net/MusicRecommendationDataset/lastfm-1K.html}}  \citep{mcfee2012million} contains listening behaviors of 992 users and we mapped the tracks jointly by artist and title into 15 genres on Discogs\footnote{\href{https://www.discogs.com/search/}{https://www.discogs.com/search/}}. We created user sequences by choosing months where a user has between 5 and 500 events, and identified distinct genres as sets for each event in a sequence, potentially resulting in multiple sequences per user. After these preprocessing steps, we obtained a total of 10,705 sequences. 

\textbf{MovieLens 25M} \citep{harper2015movielens} contains 25 million ratings for 62,000 movies from 162,000 users. We used user sequences from the year 2016.
The movies are categorized into 20 genres, and each movie can belong to multiple genres that are interpreted as items in sets. After filtering sequences to have between 5 and 200 events, we are left with 11,198 user sequences.

The \textbf{MOOC} user action dataset \citep{kumar2019predicting} represents course activities of 7,047 users with 97 different possible activities (``items''). 
Activities that have the same timestamp correspond to sets. This dataset is skewed in that only 3.7\% of events have more than a single activity (or item); this is compatible with our general framework since sets can contain a single item.
After filtering sequences to have between 5 and 200 events, we have 6,892 user sequences.

\subsection{Model Architecture and Training Details}\label{sec:exp_details_model}
Each event in the sequence is represented as a multi-hot encoding vector $\set_i = [X_{i,1}, X_{i,2},..., X_{i, K}]$ with 0/1 elements. The event vector is calculated using \cref{eq:embed} in the main paper from the embedding weights. For the temporal component, using the neural Hawkes model, we apply the Softplus function after a linear mapping of hidden states $\mathbf{h}(t)$ to obtain the total intensity $\lambda^*(t)$, where we set $s=1$ in $\lambda^*(t) = s \log ( 1 + \exp(\mathbf{u} \cdot \mathbf{h}(t) / s))$ for simplicity. In addition, we use a simple linear mapping between hidden states $\mathbf{h}(t)$ and item probabilities in \cref{eq:intensity2logits}. We also perform an ablation study using two layers for nonlinear mapping and the results are presented in \cref{sec:exp_results_ablation}. 

The hyperparemeters for model training on each of the four datasets are summarized in \cref{tab:hyperparameter} for all variants of models. We choose a larger hidden state size for MovieLens because of the very large variation in time gaps between events (from small to large) relative to the other datasets, 
and we use a larger embedding size for MOOC because of the relatively large number of items relative to the other datasets.

\begin{table}[h]
\centering
\caption{Hyperparameters for Model Architecture for Training}
\begin{tabular}{l c c}
\toprule
Dataset &  Embedding Size & Hidden State Size \\
\midrule
Instacart & 16 & 64 \\
Last.fm & 16 & 64 \\
MovieLens & 16 & 128  \\
MOOC & 32 & 64 \\
\bottomrule
\end{tabular}
\label{tab:hyperparameter}
\end{table}

In addition, we apply the following default hyperparameter settings for experiments across all model variants and all datasets. The learning rate is fixed at 0.001 with no weight decay, except for a linear warm-up learning rate being applied for the first 1\% iterations before achieving 0.001. We use a batch size of 128, and a fixed number of 300 epochs for training. The Adam stochastic gradient algorithm \citep{DBLP:journals/corr/KingmaB14} is used for optimization, where we cap the gradient at $10,000$ for stability in training.

\subsection{Discussion on DPP Variants}\label{sec:exp_details_dpp}

In our experiments with DPP, %
our implementation is based on the following expression for the likelihood: $\prob(A) = \det(\mathbf{I}_A \mathbf{K}' + \mathbf{I}_{\bar{A}} (\mathbf{I} - \mathbf{K}'))$. Here $\mathbf{K}'$ is the marginal kernel in DPPs that can be computed as $\mathbf{K}'=\mathbf{L}(\mathbf{L}+\mathbf{I})^{-1}=\mathbf{I}-(\mathbf{L}+\mathbf{I})^{-1}$, $\mathbf{I}_A$ is the diagonal matrix with ones in diagonal entries if the index corresponds to item belonging to $A$ and zeros otherwise, and $\mathbf{I}_{\bar{A}}$ has ones in diagonal entries if the index corresponds to item not in $A$. This formulation allows us to parallelize the computation with varying sizes of sets $X_i$ using standard machine learning frameworks such as PyTorch that require tensors to have the same shape on every dimension.

We did not conduct querying experiments for our DPP models for computational reasons, i.e., the standard DPP model scales
in terms of time complexity as $\mathcal{O}(K^3)$, and the eigendecomposition can also be numerically unstable as $K$ grows. 
In principle,  DPP approaches that are more computationally tractable could be pursued as alternatives within our framework.

\subsection{Empirical Distributions of $A$-before-$B$ Queries}\label{sec:exp_details_ab_query}

As described in \cref{sec:main_results_querying} in the main paper, we define an $A$-before-$B$ query to be of the form $\prob (\hit(A) < \hit(B), \hit(A) \leq t | \hist)$ conditioned on the history for each test sequence. \cref{tab:ab_distribution} summarizes the empirical distribution for the four scenarios (for each of the four datasets), showing relatively balanced distributions for each scenario given our querying setup.

\begin{table}[h]
\centering
\caption{Empirical counts of the four scenarios listed in \cref{sec:ab_query}, where we suppress the notation that the smaller value between $\hit(A)$ and $\hit(B)$ is less than or equal to $t$ unless stated otherwise.}
\begin{tabular}{l c c c c}
\toprule
Dataset & $\hit(A)=\hit(B) \leq t$ & $\hit(A) < \hit(B)$ & $\hit(B) < \hit(A)$ & $\hit(A)>t,\hit(B) > t$ \\
\midrule
Instacart & 74 & 298 & 286 & 342 \\
Last.fm & 122 & 309 & 301 & 268 \\
MovieLens & 128 & 341 &  351 & 180 \\
MOOC & 1 & 162 & 162 & 675\\
\bottomrule
\end{tabular}
\label{tab:ab_distribution}
\end{table}

\section{ADDITIONAL EXPERIMENTAL RESULTS}\label{sec:exp_results}

\subsection{Recurrent Marked Temporal Point Processes Results}\label{sec:exp_result_rmtpp}
\begin{figure}[h]
    \centering
    \includegraphics[width=0.45\columnwidth]{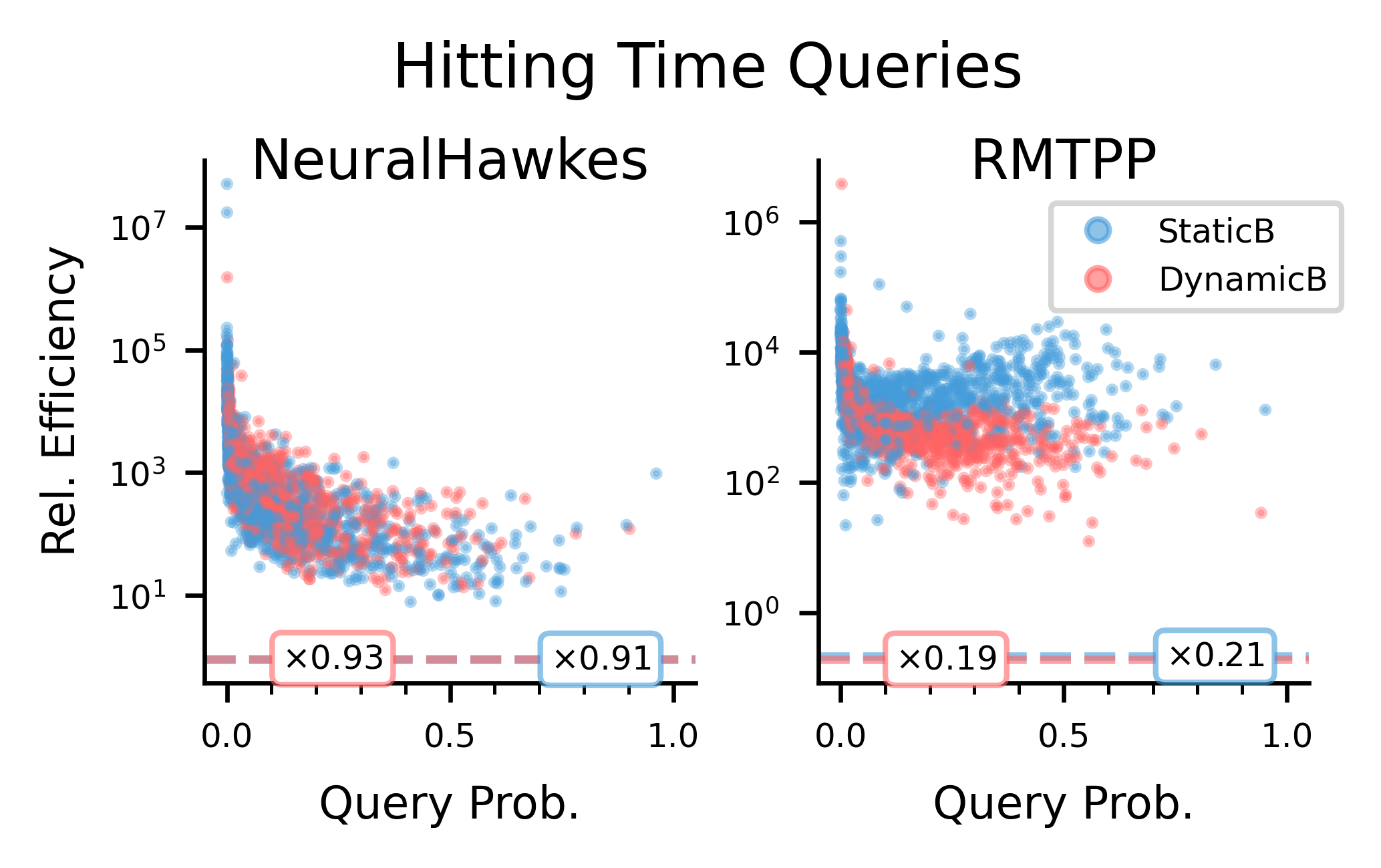}
    \includegraphics[width=0.45\columnwidth]{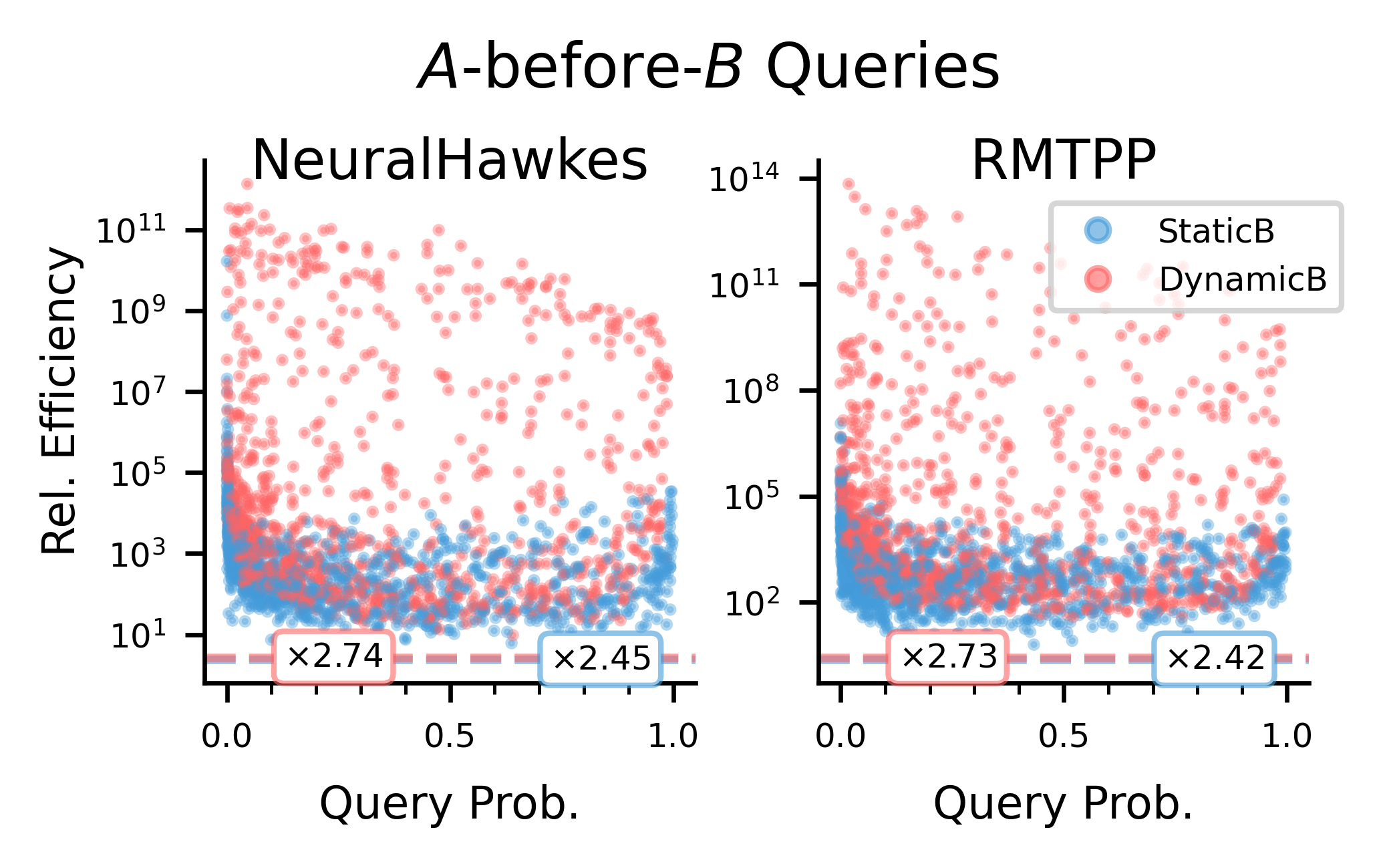}
    \caption{Relative efficiency for hitting time queries of the form $p(\hit(A) \leq t \mid \hist)$ and $A$-before-$B$ queries of the form 
    $p(\hit(A) < \hit(B), \hit(A) \leq t \mid \hist)$ 
    for two model variants with neural Hawkes and RMTPP temporal models. Blue and red dashed lines refer to the multiplicative runtime of importance sampling compared to naive sampling.}
    \label{fig:eff_rmtpp}
\end{figure}

\begin{figure}[H]
    \centering
    \includegraphics[trim={0.1cm 0.15cm 0 0},clip,width=0.45\columnwidth]{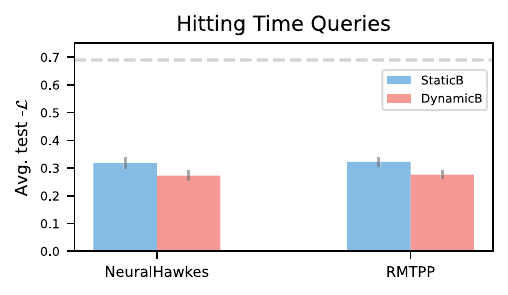}
    \includegraphics[trim={0.1cm 0.1cm 0 0},width=0.45\columnwidth]{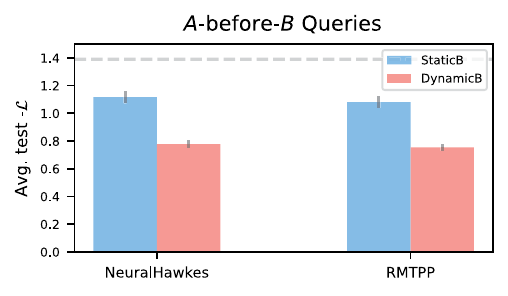}
    \caption{Average negative test log-likelihood ($\pm 1$ std. dev.) of hitting time queries $p(\hit(A) \leq t \mid \hist)$
    and  $A$-before-$B$  queries 
    $p(\hit(A) < \hit(B), \hit(A) \leq t \mid \hist)$
    estimates for two temporal models and two variants. Lower values are better: the lower bound is 0 and the upper dashed line is the negative log-likelihood of randomly guessing outcomes.}
    \label{fig:query_ll_rmtpp}
\end{figure}

To illustrate that our framework is general and compatible with any recurrent MTPP model, we investigated the use of the Recurrent Marked Temporal Point Processes (RMTPP) as an instantiation for the temporal component of our proposed model. We use the Instacart dataset as an example and directly learn the total intensity as well as the set structure from the hidden states $\mathbf{h}(t)$ in the same manner as for the neural Hawkes instantiation. Other training, sampling, and experimental details are the same as the experiments using neural Hawkes as the temporal model. \cref{tab:log_likelihood_rmtpp} presents the test sequence log-likelihood. The dynamic models systematically outperform static models, and neural Hawkes models generally learn a better temporal representation than RMTPP models. \cref{fig:eff_rmtpp} plots the relative efficiency for both temporal models for the Static and Dynamic Bernoulli variants. We observe consistent gains in sample size using importance sampling relative to naive sampling. \cref{fig:query_ll_rmtpp} shows the corresponding query log-likelihood where dynamic models are consistently superior over static models. The differences between neural Hawkes and RMTPP temporal models are not substantial.

\begin{table}[H]
\
\caption{Negative test sequence log-likelihood, $-\mathcal{L}$ from \cref{eq:logl}
, for the Instacart dataset, with different static and dynamic variants of models, comparing NH and RMTPP for the dynamic components. Also shown is the decomposition of each  $-\mathcal{L}$ into time $\mathcal{L}_\text{Time}$ and set $\mathcal{L}_\text{Set}$  components. We highlight the results for $-\mathcal{L}$  to denote the first (bold) and second (underline) best-performing models overall. 
}

\centering
\resizebox{0.5\columnwidth}{!}{ 
\begin{tabular}{p{3cm} r r r}
\toprule
Model \hspace*{\fill} &  \multicolumn{1}{c}{$-\mathcal{L} (\downarrow)$} & \multicolumn{1}{c}{$-\mathcal{L}_\text{Time}(\downarrow)$} & \multicolumn{1}{c}{$-\mathcal{L}_\text{Set}(\downarrow)$}\\
\midrule

\multicolumn{1}{l}{StaticB-Poisson} & 205.11 & 58.17 & 146.94 \\
\arrayrulecolor{gray}\midrule
\multicolumn{1}{l}{StaticB-NH} & 198.22 & 51.30 & 146.92 \\
\multicolumn{1}{l}{StaticB-RMTPP} & 203.08 & 56.15 & 146.93 \\
\gmidrule
\multicolumn{1}{l}{StaticDPP-NH} & 203.35 & 51.37 & 151.98 \\
\multicolumn{1}{l}{StaticDPP-RMTPP} & 208.36 & 56.14 & 152.22 \\
\arrayrulecolor{gray}\midrule
\multicolumn{1}{l}{DynamicB-NH} & \textbf{168.04} & 51.46 & 116.58 \\
\multicolumn{1}{l}{DynamicB-RMTPP} & 173.29 & 56.08 & 117.21 \\
\gmidrule
\multicolumn{1}{l}{DynamicDPP-NH} & \underline{170.68} & 51.41 & 119.27 \\
\multicolumn{1}{l}{DynamicDPP-RMTPP} & 175.87 & 56.08 & 119.79  \\

\arrayrulecolor{black}

\bottomrule
\end{tabular}
}
\label{tab:log_likelihood_rmtpp}
\end{table}

\subsection{Runtime Comparisons}\label{sec:exp_results_runtime}
We conducted runtime experiments on hitting time queries of the form $p(\hit(A) \leq t \mid \hist)$ for Static and Dynamic Bernoulli models coupled with neural Hawkes processes on four datasets. In these experiments, we vary the number of items in $A$ by using different percentages of items out of all possible items, while other settings are the same as the main paper. The included items are randomly chosen from existing items in the conditioned history for the MOOC dataset and from all items for the other datasets. We compare the wall-clock runtime in seconds per sample by plotting the ratio between importance sampling and naive sampling. The value indicates the multiplicative wall-clock time per sample that importance sampling takes compared to naive sampling. 

\begin{figure}[H]
    \centering
    \includegraphics[width=0.9\columnwidth]{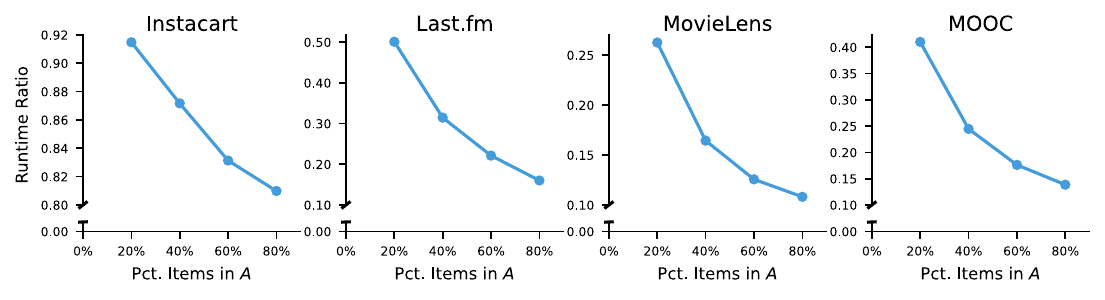}
    \caption{Runtime analysis for Static Bernoulli models. The $x$-axis refers to the percentage of items associated with the query $p(\hit(A) \leq t \mid \hist)$. The $y$-axis refers to the multiplicative increase in wall-clock time per sample of naive sampling compared to importance sampling.}
    \label{fig:hitting_runtime_static}
\end{figure}

\begin{figure}[H]
    \centering
    \includegraphics[width=0.9\columnwidth]{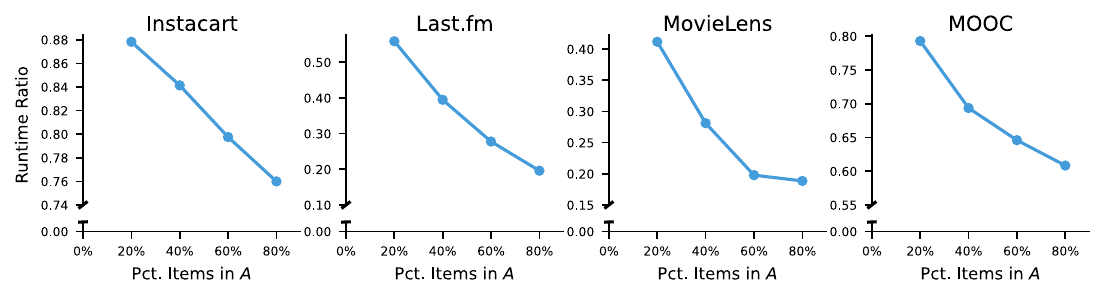}
    \caption{Runtime analysis for Dynamic Bernoulli models with the same format as \cref{fig:hitting_runtime_static}.}
    \label{fig:hitting_runtime_dynamic}
\end{figure}

\cref{fig:hitting_runtime_static,fig:hitting_runtime_dynamic} show that the ratio is consistently smaller than 1, demonstrating that importance sampling is more efficient in wall-clock runtime. We gain more efficiency in importance sampling with an increasing number of items associated with the query (approximately linear), because our proposal distribution effectively zeros out the intensities for the proportion of items included in $A$.

\subsection{Ablation Study of Dynamic Bernoulli}\label{sec:exp_results_ablation}
\begin{figure}[H]
    \centering
    \includegraphics[width=0.8\columnwidth]{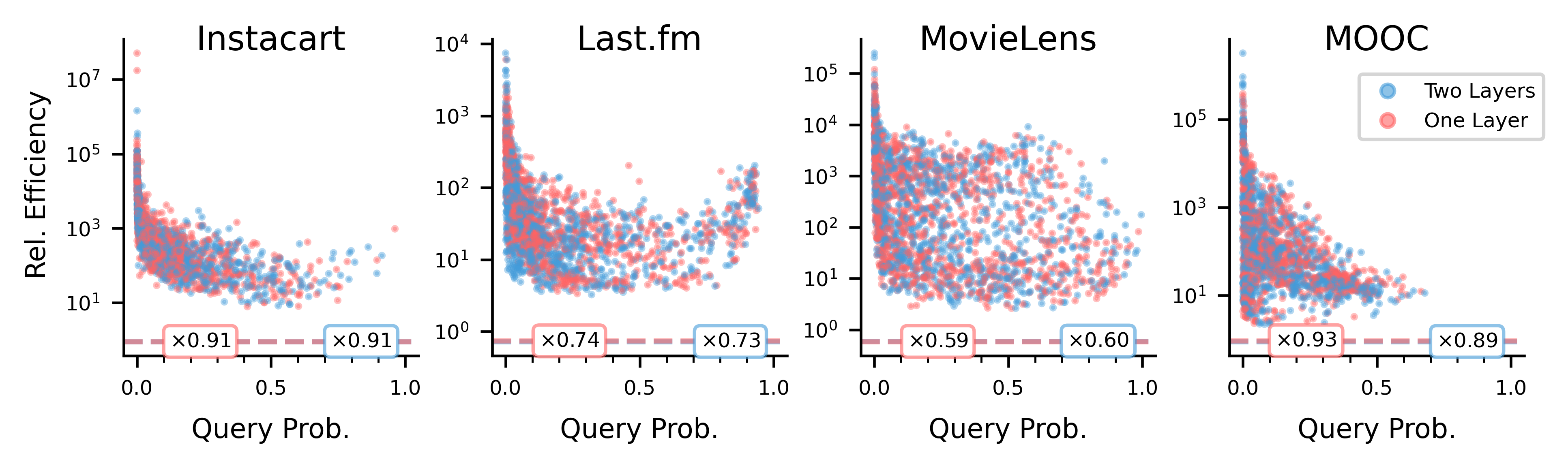}
    \caption{Relative efficiency for queries of the form $p(\hit(A) \leq t \mid \hist)$ for two model variants. Blue and red dashed lines refer to the multiplicative runtime of importance sampling compared to naive sampling.}
    \label{fig:hit_eff_2layers}
\end{figure}

\begin{figure}[H]
    \centering
    \includegraphics[width=0.8\columnwidth]{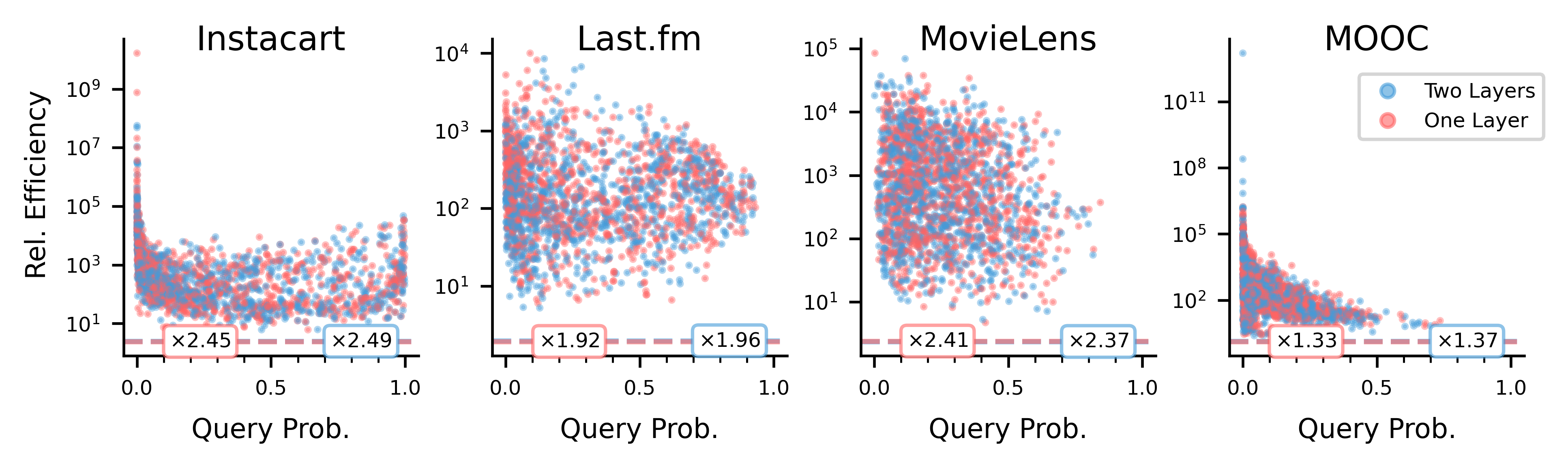}
    \caption{Relative efficiency for queries of the form 
    $p(\hit(A) < \hit(B), \hit(A) \leq t \mid \hist)$ 
    for two model variants. Blue and red dashed lines refer to the multiplicative runtime of importance sampling compared to naive sampling.}
    \label{fig:ab_eff_2layers}
\end{figure}

\begin{figure}[H]
    \centering
    \includegraphics[trim={0.1cm 0.15cm 0 0},clip,width=0.6\columnwidth]{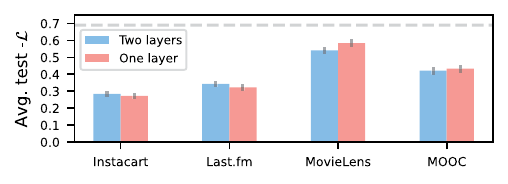}
    \includegraphics[trim={0.1cm 0.1cm 0 0},width=0.6\columnwidth]{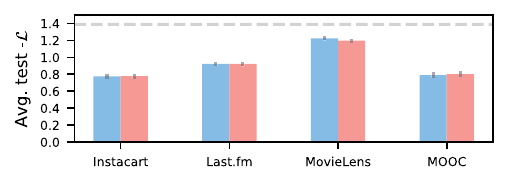}
    \caption{Comparing single-layer and 2-layer configurations for set modeling. Average negative test log-likelihood ($\pm 1$ std. dev.) of hitting time queries (top) 
    and  $A$-before-$B$  queries (bottom)  across 4 datasets.
    Lower values are better: the lower bound is 0 and the upper dashed line is the negative log-likelihood of randomly guessing outcomes.}
    \label{fig:query_ll_2layers}
\end{figure}

In all our previous experiments, we used a linear layer for $\mathbf{n}$ in $\rho_k(t) := \sigma(\mathbf{v}_k \cdot \mathbf{n}(\mathbf{h}(t)) + b_k)$ (\cref{eq:intensity2logits}). We use \textit{DynamicB-NH-2} to refer to the model using two layers in $\mathbf{n}$ for non-linear mappings. Results in \cref{tab:log_likelihood_2layer,fig:hit_eff_2layers,fig:ab_eff_2layers,fig:query_ll_2layers} show that the difference between single layer and multiple layers are not substantial.

\begin{table}[H]
\
\caption{Comparing single-layer and 2-layer (indicated as ``-2'') configurations for  set modeling. Same form as in \cref{tab:log_likelihood_rmtpp}.}

\centering
\resizebox{0.5\columnwidth}{!}{ 
\begin{tabular}{p{3cm} r r r}
\toprule
Dataset\hspace*{\fill}Model &  \multicolumn{1}{c}{$-\mathcal{L} (\downarrow)$} & \multicolumn{1}{c}{$-\mathcal{L}_\text{Time}(\downarrow)$} & \multicolumn{1}{c}{$-\mathcal{L}_\text{Set}(\downarrow)$}\\
\midrule
\multicolumn{1}{l}{\textit{Instacart}} \\
\multicolumn{1}{r}{StaticB-Poisson} & 205.11 & 58.17 & 146.94 \\
\arrayrulecolor{gray}\cmidrule(l{2em}r{0.5em}){1-4}
\multicolumn{1}{r}{StaticB-NH} & 198.22 & 51.30 & 146.92 \\
\multicolumn{1}{r}{DynamicB-NH} & 168.04 & 51.46 & 116.58 \\
\multicolumn{1}{r}{DynamicB-NH-2} & 167.93 & 51.37 & 116.56 \\
\arrayrulecolor{black}
\midrule
\multicolumn{1}{l}{\textit{Last.fm}} \\
\multicolumn{1}{r}{StaticB-Poisson} & 1027.14 & 377.17 & 649.97 \\
\arrayrulecolor{gray}\cmidrule(l{2em}r{0.5em}){1-4}
\multicolumn{1}{r}{StaticB-NH} & 415.00 & -234.91 & 649.92 \\
\multicolumn{1}{r}{DynamicB-NH} & 259.08 & -223.59 & 482.67 \\
\multicolumn{1}{r}{DynamicB-NH-2} & 262.26 & -219.28 & 481.54 \\
\arrayrulecolor{black}

\midrule
\multicolumn{1}{l}{\textit{MovieLens}} \\
\multicolumn{1}{r}{StaticB-Poisson} & 741.55 & 276.49 & 465.07 \\
\arrayrulecolor{gray}\cmidrule(l{2em}r{0.5em}){1-4}
\multicolumn{1}{r}{StaticB-NH} & 263.78  & -201.19 & 464.97 \\
\multicolumn{1}{r}{DynamicB-NH} & 236.80 & -195.78 & 432.58 \\
\multicolumn{1}{r}{DynamicB-NH-2} & 238.15 & -193.26 & 431.41 \\
\arrayrulecolor{black}

\midrule
\multicolumn{1}{l}{\textit{MOOC}}  \\
\multicolumn{1}{r}{StaticB-Poisson} & 439.74 & 169.77 & 269.97 \\
\arrayrulecolor{gray}\cmidrule(l{2em}r{0.5em}){1-4}
\multicolumn{1}{r}{StaticB-NH} & 189.54 & -81.66 & 271.20 \\
\multicolumn{1}{r}{DynamicB-NH} & 45.30 & -77.06 & 122.35 \\
\multicolumn{1}{r}{DynamicB-NH-2} & 43.99 & -75.72 & 119.71 \\
\arrayrulecolor{black}

\bottomrule
\end{tabular}
}
\label{tab:log_likelihood_2layer}
\end{table}

\subsection{Ablation Study of Hidden State Size}\label{sec:exp_results_rnn}
We empirically demonstrate that the gap of $\mathcal{L}_{\text{Time}}$ between static and dynamic models with the same RNN capacity can be reduced by increasing the size of hidden states. Using the MovieLens dataset and the neural Hawkes base model as an example, we train all models with different capacities for a fixed number of 300 epochs and report the negative log-likelihood of test sequences. Before overfitting, we observe that the gap between $\mathcal{L}_{\text{Time}}$ for dynamic and static models decreases while maintaining good performance on $\mathcal{L}_{\text{Set}}$.

\begin{table}[H]
\
\caption{Comparing different hidden state size configurations for model architecture. Same form as in \cref{tab:log_likelihood_2layer}.}
\centering
\resizebox{0.65\columnwidth}{!}{ 
\begin{tabular}{p{3cm} c r r r}
\toprule
Model(-NH) \hspace*{\fill} & \multicolumn{1}{c}{Hidden State Size} &\multicolumn{1}{c}{$-\mathcal{L} (\downarrow)$} & \multicolumn{1}{c}{$-\mathcal{L}_\text{Time}(\downarrow)$} & \multicolumn{1}{c}{$-\mathcal{L}_\text{Set}(\downarrow)$}\\
\midrule
\multicolumn{1}{l}{StaticB} & 128 & 263.78 & -201.19 & 464.97 \\
\gmidrule
\multirow{4}{*}{DynamicB} & 128 & 236.80 & -195.78 & 432.58 \\
& 256 & 231.60 & -200.94 & 432.54 \\
& 512 & 247.24 & -200.75 & 447.99 \\
& 1024 & 365.47 & -201.68 & 567.15 \\
\arrayrulecolor{gray}\midrule
\multicolumn{1}{l}{StaticDPP} & 128 & 259.95 & -203.95 & 463.90 \\
\gmidrule
\multirow{4}{*}{DynamicDPP} & 128 & 236.35 & -194.15 & 430.50 \\
& 256 & 232.93 & -199.85 & 432.78 \\
& 512 & 274.98 & -199.25 & 474.22 \\
& 1024 & 521.21 & -165.30 & 686.51 \\
\arrayrulecolor{black}

\bottomrule
\end{tabular}
}
\label{tab:log_likelihood_rnn}
\end{table}

\end{document}